%% file: arxiv.tex
\newif\ifshort
\shorttrue

\documentclass{article} 
\pdfoutput=1

\usepackage{microtype}
\usepackage[pdftex]{graphicx}
\usepackage{subfigure}
\usepackage{booktabs} 

\usepackage{hyperref}
\usepackage{natbib}

\usepackage[accepted]{icml2022}

\usepackage{amsmath}
\usepackage{amssymb}
\usepackage{mathtools}
\usepackage{amsthm}
\usepackage[algo2e]{algorithm2e}
\usepackage{placeins}
\usepackage{authblk}  
\usepackage{cleveref}
\DeclareUnicodeCharacter{2061}{}
\usepackage[flushleft]{threeparttable}
\usepackage{adjustbox}
\usepackage{multirow}
\usepackage{footmisc}
\usepackage{epstopdf}

\theoremstyle{plain}
\newtheorem{theorem}{Theorem}[section]
\newtheorem{proposition}[theorem]{Proposition}
\newtheorem{lemma}[theorem]{Lemma}
\newtheorem{corollary}[theorem]{Corollary}
\theoremstyle{definition}

\theoremstyle{remark}
\newtheorem{remark}[theorem]{Remark}
\newtheorem*{model}{Surrogate Model}

\input{math_commands.tex}

\allowdisplaybreaks
\begin{document}

\title{Rotation Invariant Quantization for Model Compression}
\author{Joseph~Kampeas, Yury~Nahshan , Hanoch Kremer, Gil Lederman, \and Shira Zaloshinski, Zheng Li and Emir Haleva 
\vskip 0.3in
 Distributed and Parallel Software Lab, Huawei Technologies\\
{\tt Email: \{first.last\}@huawei.com}
}
\date{}

\maketitle

\begin{abstract}
Post-training Neural Network (NN) model compression is an attractive approach for deploying large, memory-consuming models on devices with limited memory resources. 
In this study, we investigate the rate-distortion tradeoff for NN model compression. First, we suggest a Rotation-Invariant Quantization (RIQ) technique that utilizes a single parameter to quantize the entire NN model, yielding a different rate at each layer, i.e., mixed-precision quantization. Then, we prove that our rotation-invariant approach is optimal in terms of compression. We rigorously evaluate RIQ and demonstrate its capabilities on various models and tasks. For example, RIQ facilitates $\times 19.4$ and $\times 52.9$ compression ratios on pre-trained VGG dense and pruned models, respectively,  with $<0.4\%$ accuracy degradation.
Code is available in \href{https://github.com/ehaleva/RIQ}{github.com/ehaleva/RIQ}.
\end{abstract}


\input{sections/01_intro}

\input{sections/02_related}
\input{sections/03_prelim}

\input{sections/04_method_R}
\input{sections/05_results}
\input{sections/06_conclusion}

\FloatBarrier


\input{arxiv.bbl}
\bibliographystyle{icml2022}

\newpage
\appendix
\onecolumn

\input{sections/10_appendix}


\end{document}

%% file: math_commands.tex
\usepackage{bm,dsfont}

\def\eqref#1{equation~\ref{#1}}

\def\round#1{\left\lceil #1 \right\rfloor}
\def\1{\bm{1}}

\newcommand{\test}{\mathcal{D_{\mathrm{test}}}}

\def\eps{{ \epsilon}}
\def\teps{{\epsilon_0}}

\newcommand{\prnt}[1]{\left(#1\right)}
\newcommand{\brkt}[1]{\left[#1\right]}
\newcommand{\brcs}[1]{\left\{#1\right\}}
\newcommand{\vrnt}[1]{\left\vert #1\right\vert}
\newcommand{\inp}[2]{ \langle #1, #2 \rangle}
\newcommand{\ninp}[2]{ \langle \frac{#1}{\Vert #1 \Vert}, \frac{#2}{\Vert #2 \Vert} \rangle}

\def\rtheta{{\theta}}

\def\rw{{\textnormal{w}}}

\def\rveps{{\pmb{\epsilon}}}

\def\ww{{\pmb{w}}}

\def\rvu{{\mathbf{i}}}

\def\rvs{{\mathbf{s}}}

\def\rvu{{\mathbf{u}}}
\def\rvv{{\mathbf{v}}}
\def\rvw{{\mathbf{w}}}
\def\rvwh{{\hat{\mathbf{w}}}}
\def\rvwt{{\tilde{\mathbf{w}}}}
\def\rvx{{\mathbf{x}}}

\def\rvy{{\mathbf{y}}}

\def\rmI{{\mathbf{I}}}

\def\rmR{{\mathbf{R}}}

\def\rmU{{\mathbf{U}}}

\DeclareMathAlphabet{\mathsfit}{\encodingdefault}{\sfdefault}{m}{sl}
\SetMathAlphabet{\mathsfit}{bold}{\encodingdefault}{\sfdefault}{bx}{n}

\def\gH{{\mathcal{H}}}

\def\gM{{f}}

\def\gW{{\mathcal{W}}}

\def\ghM{{\hat{f}}} 

\newcommand{\etens}[1]{\mathsfit{#1}}

\def\etT{{\etens{T}}}

\newcommand{\E}{\mathbb{E}}

\newcommand{\R}{\mathbb{R}}



%% file: sections/01_intro.tex
\section{Introduction}

Deep neural networks are widely used for various tasks, such as computer vision, \emph{Natural Language Processing} (NLP), and recommendation systems. Nevertheless, while performance continuously improves, the models become larger with a massive increase in the number of parameters. In fact, modern \emph{Neural Network} (NN) models may have trillions of parameters, which makes the deployment of these models a challenging task \citep{chang2020train}. One way to mitigate this issue is compressing the model's parameters  to reduce its overall memory footprint while satisfying an accuracy constraint. Namely, obtaining a smaller model that is (almost) as capable as the original model. 
The most common model compression techniques are weight pruning, quantization, knowledge distillation, and low-rank decomposition. Such optimizations strive to find a smaller model while keeping the original model's accuracy, overlooking the potential inherent in its entropy limit. In the context of NN models, the entropy value is of particular interest as it provides the (theoretical) number of bits required for representing the model parameters. The optimal compression asymptotically attains this entropy limit.

In this context, lossy compression gives considerable merit as it facilitates reducing the NN size significantly with negligible accuracy degradation.  The key steps of this approach are the \emph{quantization} and the \emph{encoding} \citep[Ch. 24]{polyanskiy2022information}. In the quantization phase, the number of unique weight values (symbols) is reduced, consequently reducing the model's entropy. Modifying the symbols' statistics, however, introduces distortion (i.e., quantization error) with respect to the original model. Hence, proper quantization methods are substantial as they determine both the resulting entropy and the distortion of the model's output.  
In the encoding phase, redundant information  is removed, reaching the most compact representation possible without introducing further errors. 
Finding a solution that quantizes the model at the lowest possible bit rate while satisfying a certain distortion requirement is at the heart of quantization optimization problems, and is known as the rate-distortion problem \citep{thomas2006elements}.


Typically, NN model quantization optimizes a certain distortion (or accuracy) for a given target rate, which is suitable for reducing the computational complexity of the NN model to a lower precision. However, it is reasonable to consider the dual problem, which optimizes the compression rate while satisfying a certain distortion (or accuracy) requirement. The latter has attained very limited attention in the literature, yet, holds many fascinating challenges in its design, which motivated this study to understand the compression limits of NN. 
In this study, we investigate the rate distortion for NN models compression, where the distortion is measured by a cosine distance between the outputs of the original and the quantized model (i.e., angular deviation). Our key contribution is the thorough analysis of models rate-distortion, characterizing the minimizing distribution. In particular, we formulate the model compression as an optimization problem, where the goal is maximizing the model compression ratio subject to a given deviation constraint. Our focus is \emph{mixed-precision solutions}, where each layer gets quantized at a possibly different rate. The advantages of mixed-precision models can be manifested in NLP models, which typically have high redundancy and require extensive memory and loading times. Specifically, the main contribution is as follows.  
\begin{itemize}
\item We design a post-training Rotation-Invariant Quantization (RIQ) which compresses NN models subject to a deviation constraint. The main theme of our approach is picking the quantization bin width to be proportional to the layers' norm. Since both norm and cosine distance are invariant to rotations, this yields the optimal solution in terms of rate distortion.  To find the optimal solution efficiently, we first derive the scale in which the rate increases with the deviation and then suggest a searching paradigm that leverages our finding.   
\item To analyze the optimality of the RIQ algorithm, we introduce a universal surrogate model that depicts quantization in terms of rotation of the model weights. Its analysis reveals that the rate-distortion minimizing distribution for NN models is a spherical (rotation invariant) distribution constructed by the product of layers' spherical distribution. Due to convexity, the rate achieved under this product distribution is bounded by a rate achieved under the layers' average spherical distribution. Since RIQ optimizes the rate only over spherical distributions, it reaches the optimal solution efficiently. 
\item We rigorously evaluate the RIQ and demonstrate its capabilities on various models and tasks. RIQ attains a remarkable compression ratio with a negligible accuracy loss in all assessments, surpassing recent results in the literature. 
\end{itemize}

%% file: sections/02_related.tex
\section{Related Work}
This section is devoted to prior work on model compression that is most relevant to this study. Roughly speaking, typical model compression methods can be classified into four categories. Weight pruning, quantization, knowledge distillation \citep{sarfraz2021knowledge, walawalkar2020online}, and low-rank decomposition \citep{idelbayev2020low,lin2018holistic,lee2019learning}. Even though such methods strive to find a smaller model while retaining the model's accuracy, they often tend to neglect the potential inherent in the entropy limit. In this study, we seek to minimize the model entropy by quantization and then attain this entropy limit by compression while satisfying a distortion requirement at the model's output.   

Quantization is a prominent method for compressing NN models. In \citet{wu2020integer, banner2019post, idelbayev2021optimal}, the authors considered fixed-bit quantization methods, where all layers are quantized at the same integer bit rate. In this paper, on the other hand, we consider post-training mixed-precision solutions. \citet{Bhalgat_2020_CVPR_LSQplus, wang2019haq} focused on quantization-aware training, where the weights quantization is performed during the training. To attain lower quantization rates, \citet{angela2020quantnoisetrain,baskin2021uniq, defossez2021differentiable} suggested training the models with noise. Although  quantization-aware training methods may achieve better results than the post-training approaches, they are time-consuming, and requires ample representative data and training resources. 
Nevertheless, these works optimize the distortion, while RIQ optimizes the rate for a given distortion requirement. 

The idea of pruning NN connections based on information-theoretic ideas was explored already in the seminal works \citep{lecun1989optimal,hassibi1993optimal}. Later, \citet{songhan2015deepcompression} used magnitude threshold to remove redundant parameters and then utilized Huffman's entropy coding to compress these parameters. Since then, pruning techniques gained popularity, searching for effective methods to prune parameters \citep{zhang2021exploration, frankle2018lottery, lee2019learning},  showing that entropy reduction during training is beneficial, as low-entropy models are more amenable to compression  \citep{oktay2019scalable, Baskin_Cat}. Yet, the first rate-distortion formulation of model compression was suggested by \citet{gao2019rate,isik2022information}, providing significant insights about the optimal model pruning. In particular, \citet{gao2019rate} studied the two-layer networks, providing tractable analysis when assuming the Gaussian distribution of the model weights. This assumption, however, may not hold in general. \citet{isik2022information} observed empirically that the weights of a pretrained model tend to follow the Laplacian distribution, for which the optimal compression algorithm that resides on the rate-distortion curve must output a sparse model. This study analyzes the rate-distortion of rotation-invariant quantization solutions, which generalizes these findings, as the Gaussian and Laplacian distributions are rotation-invariant distributions. 

%% file: sections/03_prelim.tex
\section{Preliminaries}


In this section, we rigorously define the model compression optimization problem and the relevant known results on quantization and the rate-distortion theory. Throughout, bold $\rvw$ denote weight vectors, unless stated otherwise. $\Vert \cdot \Vert$ and $\inp{\cdot}{\cdot}$ denotes the standard $\ell^2$-norm and the inner product, respectively. Further, the model weights are treated as random variables and we use $p_\rw(w)$ to denote the probability distribution of a random variable $\rw$. In other words, due to the random seed from which the weights are initialized, the resulting weights after the training can be considered as a function of this random seed. Clearly, initializing the weights with a different seed would yield a different realization of the models' weights.  Hereafter, $\rvw_{[1:L]} = \brcs{\rvw_{1}, ... ,\rvw_{L}} \in \R^{N}$, denotes the concatenation of the weights of a pretrained model with $L$ layers, where $\rvw_{\ell} \in \R^{n_\ell}$ are $n_\ell$ weights of layer $\ell$ and $N = \sum_{\ell=1}^L n_\ell$. The quantized representation of those weights is denoted by $\rvwh_\ell$.

\subsection{Problem Statement}
Let $\gM: \R^{n_{x}} \to \R^{n_y}$ be a \emph{pretrained model} that characterizes the prediction of input $\rvx$ to an output $\rvy \in \R^{n_y}$. The model comprises $L$ intermediate mappings called layers, such that $\gM(\rvx) = \gM_{\rvw_L}\prnt{\gM_{\rvw_{L-1}}\prnt{\cdots \gM_{\rvw_1}(\rvx)}}$, where $\rvw_\ell \in \R^{n_\ell}$ are the weights of layer $\ell$. We further assume that each layer performs an affine operation (e.g., convolution) followed by a nonlinear operation (e.g., ReLU).   

Our goal is to obtain the smallest (quantized and compressed) version of this model $\ghM$, whose output is as close as possible to the output of $\gM$. To assess the fidelity of the quantization, a sample $\rvx$ is sent through $\gM$  and $\ghM$, and the  \emph{deviation between the outputs} $\gM(\rvx)$ and $\ghM(\rvx)$ is measured. In this study, we focus on the cosine distance as distortion measure. That is, 
\begin{equation}
    d_{\gM, \rvx}(\rvw_{[1:L]},\rvwh_{[1:L]}) \triangleq 1 - \frac{\inp{\gM(\rvx)}{\ghM(\rvx)}}{\Vert \gM(\rvx)\Vert \cdot \Vert \ghM(\rvx) \Vert}
    \label{eq: cosine dist def}
\end{equation}
This distortion reflects the rotation angle that is required to align $\ghM(\rvx)$ with $\gM(\rvx)$, including both the models activation and non-linearity. Noticeably, \cref{eq: cosine dist def} is a rotation-invariant measure, which means that its value does not change when an arbitrary rotation is applied to its arguments. That is, the cosine value does not change when rotating together $\ghM(\rvx)$ and $\gM(\rvx)$. 
Other distortion measures, such as the $l_2$ and $l_1$ distances, can also be considered for analyzing model compression \citep{gao2019rate, isik2022information}. However, due to the curse of dimensionality, they are mainly useful when addressing low-dimensional optimization \citep{aggarwal2001surprising}.  On the other hand, cosine distance is beneficial to measure the similarity between vectors in high-dimensional spaces, which naturally occurs in models. Further, the correctness of the models in many tasks is determined by the orientation of the output vector rather than the magnitude of the output vector. This motivates us to optimize the quantization rate under the cosine distance measure. Hereafter, the term \emph{deviation} is used to depict $d_{\gM, \rvx}(\rvw_{[1:L]},\rvwh_{[1:L]})$, which is the cosine distance between the outputs, whereas the term \emph{distortion}  refers to the cosine distance $d(\rvw_\ell, \rvwh_\ell)$ between the weights $\rvw_\ell$ and their quantized representation $\rvwh_\ell$. 
 


As $\rvx$ passes the first layer of $\gM$ and $\ghM$, it is rotated (and scaled) by $\rvw_1$ and $\hat{\rvw}_1$, respectively. Due to the quantization\ifshort, \fi $\rvw_1$ and $\hat{\rvw}_1$ acts differently on $\rvx$, and thus, yields unequal outputs. These unequal outputs are rotated (and scaled) by the next layer's weights, and so on, reaching the output of $\gM$ and $\ghM$. The resulting deviation in \cref{eq: cosine dist def} relates to the distortions gathered through the layers. In particular, each quantized layer produces a rotation distortion in its output, and this distortion keeps propagating and accumulating through the layers until reaching the model's output.

Thus, characterizing the connection between the deviation to the distortion in each layer is a key to optimizing the model quantization. Though finding the exact link is intricate in general, intuitively, as  the layers' quantization rate jointly decrease,  the deviation increases monotonically with the layers' distortion. Since these distortions are invariant to rotations, the rate-distortion theory tells that the optimal quantization must be rotation invariant as well, as we show in the sequel.  
This motivates a searching paradigm over rotation-invariant solutions, where the layers' rate are jointly adjusted.     

Formally, given a trained model $\gM$, and a sample $\rvx$, we wish to find a quantized model $\ghM$ whose weights $\hat{\rvw}_{[1:L]}$ solves the following optimization problem. 
\begin{equation*}
\begin{aligned}
& \min 
& & \text{Rate} (\hat{\rvw}_{[1:L]}) \\
& \text{s.t.}
& & d_{f,x}(\rvw_{[1:L]},\rvwh_{[1:L]}) \leq D
\end{aligned}
\label{eq: optimization def}
\end{equation*}
for some deviation requirement $D$, where rate is the average bits per symbol over the entire quantized model $\ghM$. 
In this work, we characterize the properties of the minimum rate and devise a searching method that finds the minimum rate that satisfies $D$ efficiently. 

In this formulation, we consider \emph{mixed-precision quantization} solutions, where the weights $\rvw_\ell$ of each layer $\ell$ are quantized at a different rate $R_\ell$. 
In other words, each layer utilizes a different number of symbols. Typically, using fewer symbols induces lower entropy, and hence, we strive to minimize the number of symbols in each layer. Though the entropy indicates the shortest representation of a layer, one must encode the layers' weights to reach this entropy limit. In this study, we utilize \emph{Asymmetric Numeral Systems} (ANS) arithmetic encoder for this purpose \citep{duda2013asymmetric}. 
Accordingly, the resulting compression ratio, assuming 32 bits representation of the source symbols, is   
\begin{equation}
  \text{Weights Compression Ratio} \approx \frac{32 \cdot \sum_{\ell=1}^L n_\ell}{\sum_{\ell=1}^L n_\ell \cdot H(\rvwh_\ell) + \vert T_\ell \vert}
\label{eq:cmp_ratio}
\end{equation}
Where $\vert T_\ell \vert$ denotes the coding table size of layer $\ell$ and $H(\rvwh_\ell)$ is the empirical entropy. 

\subsection{Rate-Distortion Theory}
The rate-distortion theory determines the \emph{minimum number of bits per symbol}, or simply the minimum bit rate,  required for describing a random variable with a certain  (average) distortion.  In particular,  to quantize  a  sequence of $n$ independent realizations $\pmb{w} = (w_1,\cdots, w_n)$, generated by a source $\rvw$ with distribution $p_\rvw(\pmb{w}), \pmb{w}\in \gW^n$ into $R$ bits,  encoding and decoding functions are utilized. The encoder $\mathcal{E}: \gW^n \to \brcs{0,1}^{R}$ maps the sequence to one of $2^{R}$ possible indices, and the decoder  $\mathcal{D}: \brcs{0,1}^{R} \to \hat{\gW}^n$ maps the given index into an estimated (quantized) sequence $\hat{\pmb{w}} = (\hat{w}_1,\cdots, \hat{w}_n)$. Thus, the rate-distortion pair $(R,D)$ are the resulting rate $R$ and distance $D= d(\pmb{w}, \hat{\pmb{w}})$ between the original sequence and the quantized sequence. 

In general, we wish to minimize both the rate and the distortion, however, there is an inherent tradeoff, characterized by rate-distortion function as \citep[Ch. 10]{thomas2006elements} 
\begin{align}
R(D) &= \min_{p(\hat{\pmb{w}} | \pmb{w}):\E\brkt{ d(\pmb{w}, \hat{\pmb{w}}))} \leq D} I(\rvw;\rvwh)
\label{eq: rate distortion def}
\end{align}
where $I(\rvw;\hat{\rvw}) = H(\rvw) - H(\rvw | \rvwh)$ is the mutual information between the source vector $\rvw$ and its reconstruction $\hat{\rvw}$ \citep[Ch. 2.4]{thomas2006elements}, and $d(\cdot, \cdot)$ is a predefined distortion metric, such as the cosine distance in \cref{eq: cosine dist def}. Thus, the rate-distortion function determines the infimum  rate $R$ that achieves a given distortion $D$. This infimum is attained by minimizing overall conditional distributions $p(\hat{\pmb{w}} | \pmb{w})$ for which distortion $D$ is satisfied under $p(\pmb{w})$. 



\subsection{Uniform Scalar Quantization}\label{sec. scalar quant}
The rate-distortion theory tells that it is optimal to describe the whole sequence jointly, using one of $2^{R}$ indices, even when the variables are i.i.d. Yet, in terms of entropy, \citet{koshelev1963quantization} showed that uniform scalar quantization is (asymptotically) optimal when one intends to further compress the quantized data losslessly. Since this paper considers the latter approach,  this section briefly discusses uniform scalar quantization and its analysis.  

For a random variable $\rw \in \brkt{-A/2,A/2}$, where $A \in \R$, uniform quantization partitions the range $\brkt{-A/2,A/2}$ into $N$ bins uniformly, such that each bin has width $\Delta = A/N$.  
Thus, any realization of $\rw$ is encoded (rounded) into \emph{an integer value}, $\round{\rw/\Delta}$, that corresponds to its bin index.  The decoder then reconstructs its value by
\begin{equation}
  \hat{\rw} = \round{{\rw}/{\Delta}} \cdot \Delta
  \label{eq: reconstruct w}
\end{equation}

The fidelity of this quantization is measured by a distortion measure, such as the \emph{Mean Squared Error} (MSE) criterion, defined as $D(N)=\E\vert \rw-\hat{\rw}\vert^2$. To analyze, it is more convenient to examine the quantization in terms of rate  $R=\log_2 ⁡N$. In high-rate regime (i.e., $R \gg 1$), the \emph{probability density} in each bin is nearly flat, and hence, the distortion is \citep[Ch. 24.1]{polyanskiy2022information} 
\begin{equation}
    \E \vert \rw- \hat{\rw} \vert^2  = \Delta^2/12
     \label{eq. scalar quant}
\end{equation}
Further, the resulting entropy of the quantized symbol is \citep[Theorem 8.3.1]{thomas2006elements} 
\begin{equation}
    H(\hat{\rw}) = h(\rw) - \log(\Delta) \text{\quad     [bits/symbol]}
    \label{eq: entropy gain}
\end{equation}
where $h(\rw) = -\int p(w) \log p(w)\mathrm{d}w$ is the differential entropy function \citep[eq. (8.1)]{thomas2006elements}. 
In other words, quantization reduces the entropy by $\log \Delta$, and thus, a larger $\Delta$ yields a lower entropy, and hence, potentially, a higher compression ratio by \cref{eq:cmp_ratio}. 

%% file: system_model_tikz.tex
\begin{figure}
    \centering
\begin{tikzpicture}[font=\normalsize,thick, scale=0.7, transform shape]
 
\node[draw,
    rounded rectangle,
    minimum width=2.5cm,
    minimum height=1cm] (block1) {Input $\rvx$};
    %

\node[draw,
    below left=of block1,
    minimum width=2.cm,
    minimum height=1cm
] (w1) { $\sigma\prnt{\rvw_1 (\cdot)}$};

\node[draw,
    below=of w1,
    minimum width=2.cm,
    minimum height=1cm
] (w2) { $\sigma\prnt{\rvw_2(\cdot)}$};

\node[draw,
    below=2.cm of w2,
    minimum width=2.cm,
    minimum height=1cm
] (wL) {$\sigma\prnt{\rvw_L (\cdot)}$};

\node[draw,
    below right=of block1,
    minimum width=2.cm,
    minimum height=1cm
] (qw1) { $\sigma\prnt{\hat{\rvw}_1(\cdot)}$};

\node[draw,
    below=of qw1,
    minimum width=2.cm,
    minimum height=1cm
] (qw2) { $\sigma\prnt{\hat{\rvw}_2(\cdot)}$};

\node[draw,
    below= 2.cm of qw2 ,
    minimum width=2.cm,
    minimum height=1cm
] (qwL)  {$\sigma\prnt{\hat{\rvw}_L(\cdot)}$};

\foreach \y in {0,0.5,1}
    \draw (w2) +(0,-1-\y)  circle (0.07cm) [fill=black]
    (qw2) +(0,-1-\y)  circle (0.07cm) [fill=black];

\node[draw,
    diamond,
    below=8.5cm of block1,
    minimum width=2.5cm,
    minimum height=1.5cm
    ] (end) { $\epsilon$};

\draw[-latex] (block1) edge (w1)
    (block1) edge (qw1)
    (w1) edge (w2)  
    (qw1) edge (qw2) 
    (wL) edge (end)
    (qwL) edge (end);


\node[ellipse,
    draw,
    above left = 1.cm and -0.7cm of qw1,
    minimum width=1cm,
    minimum height=0.7cm,
    rotate=60,
    fill = lightgray] (qc1) {};

\draw[red,-latex]  (qc1.center) -> (qc1.east) ;
\draw[blue,-latex] (qc1.center) -> (qc1.north);    

\node[ellipse,
    draw,
    above right = 0.35cm and -0.1cm of w1,
    minimum width=1cm,
    minimum height=0.7cm,
    rotate=60,
    fill = lightgray] (c1) {};

\draw[red,-latex]  (c1.center) -> (c1.east) ;
\draw[blue,-latex] (c1.center) -> (c1.north);    

 \node[ellipse,
    draw,
    below left= 0.5cm of w1,
    minimum width=1.cm,
    minimum height=0.7cm,
    rotate=-26,
    fill = cyan!20] (e1) {};
    
\draw[red,-latex]  (e1.center) -> (e1.east) ;
\draw[blue,-latex] (e1.center) -> (e1.north);

 \node[ellipse,
    draw,
    below right= 0.8cm of qw1,
    minimum width=1.cm,
    minimum height=0.7cm,
    rotate=42,
    fill = yellow!20] (qe1) {};
    
\draw[red,-latex]  (qe1.center) -> (qe1.east) ;
\draw[blue,-latex] (qe1.center) -> (qe1.north);
 
 \node[ellipse,
    draw,
    below left= 1.5cm and 0.5cm of w2,
    minimum width=1.cm,
    minimum height=0.7cm,
    rotate=-70,
    fill = cyan!20] (e2) {};
    
\draw[red,-latex]  (e2.center) -> (e2.east) ;
\draw[blue,-latex] (e2.center) -> (e2.north);

 \node[ellipse,
    draw,
    below right= 0.7cm and 0.75cm of qw2,
    minimum width=1.cm,
    minimum height=0.7cm,
    rotate=-50,
    fill = yellow!20] (qe2) {};
    
\draw[red,-latex]  (qe2.center) -> (qe2.east) ;
\draw[blue,-latex] (qe2.center) -> (qe2.north);
 
 \node[ellipse,
    draw,
    below right= 0.8cm and -0.4cm of wL,
    minimum width=1.cm,
    minimum height=0.7cm,
    rotate=-20,
    fill = cyan!20] (eL) {};
    
\draw[red,-latex]  (eL.center) -> (eL.east) ;
\draw[blue,-latex] (eL.center) -> (eL.north);

 \node[ellipse,
    draw,
    below left= 1.0cm and -0.6cm of qwL,
    minimum width=1.cm,
    minimum height=0.7cm,
    rotate=-10,
    fill = yellow!20] (qeL) {};
    
\draw[red,-latex]  (qeL.center) -> (qeL.east) ;
\draw[blue,-latex] (qeL.center) -> (qeL.north);

\node[circle,
    draw,
    below = 2.5cm  of block1,
    minimum width=0.7cm,
    minimum height=0.7cm,
    rotate=-68,
    fill = red!10] (err1) {} ;

\draw[red,-latex]  (err1.center) -> (err1.east) ;
\draw[blue,-latex] (err1.center) -> (err1.north);    

\node[circle,
    draw,
    below = 5.cm  of block1,
    minimum width=0.7cm,
    minimum height=0.7cm,
    rotate=-20,
    fill = red!10] (err2) {} ;

\draw[red,-latex]  (err2.center) -> (err2.east) ;
\draw[blue,-latex] (err2.center) -> (err2.north);    

\node[circle,
    draw,
    minimum width=1cm,
    minimum height=1cm,
    rotate=10,
    fill = red!10] (errL) at (end)  {\huge $\mathbf{\epsilon}$} ;
    
\draw[red!30,-latex]  (errL.center) -> (errL.north east) ;
\draw[blue!30,-latex] (errL.center) -> (errL.north west); 
\end{tikzpicture}
   \caption{Quantization optimization flow. $\gM$ and $\ghM$ are fed with the same input sample $\rvx$. Naturally, $\ghM$ has rotation shifts in each layer with respect to $\gM$ that propagates over the model until reaching the output. Among all quantized models $\ghM$ that satisfies the cosine distance requirement, we wish to pick the one with the smallest entropy.}
   \label{fig:sys_model}
\end{figure}

%% file: sections/04_method_R.tex
\section{Rotation-invariant  Mixed-Precision Quantization}\label{sec. frac}
In this section, we present the RIQ method, which yields a different quantization rate in each layer (i.e., mixed-precision solution) while satisfying the deviation requirement in \cref{eq: cosine dist def}. Then, we use the rate-distortion theory to analyze its performance. 

\subsection{The RIQ Algorithm}
Given a model and a deviation requirement, it is sufficient to optimize the rate over rotation-invariant solutions. 
Typical quantization methods which optimize the distortion for a given rate, on the other hand, determine $\Delta_\ell$ according to the range  $\max(\rvw_\ell) - \min(\rvw_\ell)$, whose value depends on the orientation of $\rvw_\ell$, and hence, are not rotation invariant, which is substantial for optimality. 

RIQ designs the bin width, $\Delta_\ell$, in proportion to the norm $\Vert \rvw_\ell \Vert$ in each layer. Since norm is invariant to rotations, the resulting $\Delta_\ell$ is indifferent to the orientation of $\rvw_\ell$. Consequently, the resulting distortion is also indifferent to the orientation of $\rvw_\ell$, as the bin width dictates the distortion. 

Let $\theta_\ell$ be the rotation angle from $\rvw_\ell$ to $\rvwh_\ell$ such that  $\ninp{\rvw_\ell}{\rvwh_\ell} \triangleq \cos(\theta_\ell)$. The following lemma examines the relation between $\Delta_\ell$, the norm $\Vert \rvw_\ell \Vert$, and the distortion $d(\rvw_\ell, \rvwh_\ell) = 1 - \cos(\theta_\ell)$ in the high-rate region. Hereafter, high-rate region is defined as the rate region for which $\Vert \rvwh_\ell \Vert = \Vert \rvw_\ell \Vert + o(\Vert \rvw_\ell \Vert )$. This translates to $R \geq \log^c(\Vert \rvw_\ell \Vert)$ for $c>1$. 

\begin{lemma}
Let $\eps_\ell \triangleq 1 - \cos(\theta_\ell)$ be the distortion of layer $\ell$. Then, in the high-rate region, the quantization bin width asymptotically satisfies
\begin{align*}
   \Delta_\ell &= \sqrt{\eps_\ell} \cdot \Vert \rvw_\ell \Vert \cdot \sqrt{24/ n_\ell}
\end{align*}
\vspace{-\baselineskip}
\label{coro: delta to eps}
\end{lemma}
\vspace{-1.2\baselineskip}
\ifshort
The proof of Lemma~\ref{coro: delta to eps} is elaborated in \Cref{sec. proof of coro1}.
\fi
Lemma~\ref{coro: delta to eps} links between angular distortion, bin width, norm, and dimension. For instance, from the lemma follows that  $1-\cos(\theta_\ell) =   \frac{n_\ell \cdot \Delta_\ell^2} {24\cdot \Vert \rvw_\ell \Vert^2 }$, i.e., the distortion scales linearly with the dimension $n_\ell$. Interestingly, this connection expands to the entire model as follows.
\begin{corollary}
Let $\rvw_{[1:L]}$  be a vector representation of the weights and $\rvwh_{[1:L]}$ denotes its quantized representation, and let $\theta_{[1:L]}$ be the rotation angle from $\rvw_{[1:L]}$ to $\rvwh_{[1:L]}$ such that  $\ninp{\rvw_{[1:L]}}{\rvwh_{[1:L]}} \triangleq \cos(\theta_{[1:L]})$. Considering the high-rate region, where $\Vert \rvwh_\ell \Vert = \Vert \rvw_\ell \Vert + o(\Vert \rvw_\ell \Vert)$, then, 
\[
 \cos(\theta_{[1:L]}) = \sum_{\ell=1}^L \frac{\Vert \rvw_\ell \Vert^2}{\Vert \rvw_{[1:L]} \Vert^2} \cos(\theta_\ell)  +o\prnt{\frac{\Vert \rvw_\ell \Vert^2}{\Vert \rvw_{[1:L]} \Vert^2}}
\]
\label{coro: distortion to parameters distortion}
\end{corollary}
\vspace{-\baselineskip}
The proof is deferred to  \Cref{sec. proof of coro2}. 
In words, a rotation of $\rvw_{[1:L]}$ translates to a convex combination of the layers' rotation, and vice versa. Interestingly, due to convexity of the rate-distortion, it is beneficial to average over as many rotations as possible, which means considering partitioning the model parameters into shorter vectors as we show in the sequel. Still, the most natural partition of the model is simply to it layers, which is considered herein.    

The connection of $\Delta_\ell$ to $\Vert \rvw_\ell \Vert$ in Lemma~\ref{coro: delta to eps} hints at the rotation-invariant nature of the optimization. To focus on rotation-invariant solutions, RIQ introduces a search parameter $k$ that maintains proportion with $\Vert \rvw_\ell \Vert$, allowing efficient search over these solutions.  Specifically, when $\Delta_\ell(k) = \Vert \rvw_\ell \Vert/k$  where $k$ to be optimized, the bin-width is indifferent to the orientation of $\rvw_\ell$ (i.e., rotation invariant). Further, letting the bin width grow linearly with $\Vert \rvw_\ell \Vert$ results in distortion $\eps_\ell  =  1-\cos(\theta_\ell) = \frac{n_\ell} {24\cdot k^2 }$, and hence by Corollary~\ref{coro: distortion to parameters distortion},  $1-\cos(\theta_{[1:L]}) =   \frac{1}{24\cdot k^2 }\sum_{\ell = 1}^L \frac{n_\ell \cdot \Vert \rvw_{\ell} \Vert^2}{\Vert \rvw_{[1:L]} \Vert^2} $. Namely, the distortion of the parameters scales as $O(1/k^2)$. Remarkably,  when including activation and non-linearity of models, the deviation still scales as $O(1/k^2)$. 
\begin{proposition}
In the high-rate region, the deviation in \cref{eq: cosine dist def} under RIQ scales as $O(1/k^2)$. 
\label{prop: monotonically decreasing}
\end{proposition}
The proof is deferred to  \Cref{sec. proof of prop3}. 
Essentially, since $\Delta_\ell(k)$ and $\eps_\ell$ are monotonically decreasing with $k$, then by \cref{eq: entropy gain}, the entropy increases with $k$. This allows RIQ is to reach the smallest $k$ solution (i.e., minimum entropy) which satisfies the deviation requirement. 

Next, we introduce an efficient iterative searching algorithm for finding the optimal $k$. In each iteration, the algorithm refines the searching range until reaching the smallest $k$ (up to a small constant) that satisfies the deviation requirement.  
Practically, however, as $k$ increases, $\Delta_\ell(k) \to 0$. To prevent this, we  add a small constant $\teps$ to $\sqrt{\eps_\ell}$, which bounds the value of the smallest $\Delta_\ell (k)$. In this case, setting  $\sqrt{\eps_\ell} = \frac{1}{k} \sqrt{\frac{n_\ell}{24}} + \teps$, yields,  
\begin{align}
    \Delta_\ell(k) & = \ifshort \Vert \rvw_\ell \Vert\cdot \prnt{\frac{1}{k} + \teps \cdot \sqrt{\frac{24  }{n_\ell}}} & \fi
    \label{eq: scale}
\end{align}



Even though the search of $k$ is unbounded in general, practically it is sufficient to search in bounded space since the weights' norm is finite \citep{idelbayev2021optimal}.
In the following proposition, we derive searching bounds for the optimal $k$. Let $k^*$ be the optimal (smallest) $k$ that satisfies constraint $D$, and let $\ell^*$ be the index of the layer with the largest $n_\ell$ in $\gM$. 
\begin{proposition}
The optimal $k^*$ satisfies the following bounds:
$ \frac{\sqrt{n_{\ell^*} / 24}}{(1-\teps)} \leq k^* \leq \frac{\sqrt{n_{\ell^*} / 24}}{(\teps \cdot \sqrt{\teps})}$.
\label{prop. k bounds}
\end{proposition}
\ifshort
The proof is deferred to  \Cref{sec. proof of prop2}.
\fi
To further improve the search time, a nested refinement is utilized. Specifically,
at each stage, only $\sqrt{\vrnt{O(k)}}$ values of $k$ in ascending order are considered. Once a certain value of $k$ satisfies the deviation requirement $D$, this $k$ becomes the new upper limit for the search, and the search region is refined within a smaller region of $k$ to consider, again with only $\sqrt{\vrnt{O(k)}}$ potential values to inspect. This repeatedly continues until the search step is sufficiently small (e.g., step $\leq 3$), and the compression gain becomes negligible. These refinements enable fast convergence in relatively few iterations. See Algorithm~\ref{alg: rotation inv} for a detailed description. 

\begin{algorithm}[t]
\small
\caption{The RIQ algorithm}\label{alg: rotation inv}
\KwData{model weights $\rvw_{[1:L]}$, distortion requirement $D$, minimum error $\teps$}
\KwResult{quantized weights $\rvwh_{[1:L]}$, such that $d_{\gM, \rvx}(\rvw_{[1:L]},\rvwh_{[1:L]}) \leq D$}
\textbf{Initialize:} $k_{\min} = \frac{\sqrt{n_{\ell^*}/24}}{1- \teps}$ ,  $\quad k_{\max} = \frac{\sqrt{n_{\ell^*}/24}}{\sqrt{\teps}\cdot\teps}$,
 $\quad k = k_{\min}$, $\quad \textnormal{step} = \sqrt{k_{\max} - k_{\min}}$\;
 \While{$k \leq k_{\max}$}{
    \For{$\ell = 1, \cdots, L$}{
        $\Delta_\ell = \Vert \rvw_\ell \Vert\cdot \prnt{\frac{1}{k} + \teps\cdot\sqrt{\frac{24}{n_\ell}}}$\; and quantize by:        
        $\rvwh_\ell = \round{\frac{\rvw_\ell}{\Delta_\ell}}\cdot \Delta_\ell$ \; 
        }
         
  \eIf{$d_{\gM, \rvx}(\rvw_{[1:L]},\rvwh_{[1:L]}) \leq D$}
  { 
    \eIf{$\textnormal{step} \leq 3 \quad$(stop condition) }
    {compress to $H(\rvwh_{[1:L]})$ with entropy achieving encoder\;}
    {$k_{\max} = k$\;  
    $\textnormal{step} = \sqrt{\textnormal{step}}$\;
    $k = k - \textnormal{step}\cdot \left\lfloor \textnormal{step} \right\rfloor$}
  }{
    $k = k + \textnormal{step}$
  }
}
\label{algorithm}
\normalsize
\end{algorithm}
%

\begin{remark}\label{remark:rate enforcement}
The additional degree of freedom that $\teps$  gives is substantial. For example, it facilitates enforcing quantization to a maximum of $R$ bits  (e.g., $R=8$  bits) for low precision runtime, by setting the limit $k \to \infty$ at
\ifshort
$ \teps(\ell) = \frac{\max(\rvw_\ell) - \min(\rvw_\ell)}{2^R-1} / \sqrt{\frac{24 \cdot \Vert \rvw_\ell \Vert^2 }{n_\ell}}$.
\fi
\end{remark}
For simplicity, in the sequel we apply the same small common constant value $\teps$ to all layers.

\subsection{RIQ Rate-Distortion Analysis}\label{sec. rd analysis}
In this section, we provide theoretical justification for the optimality of RIQ. We introduce a surrogate model for which the rate-distortion analysis with cosine distance is tractable, showing that the minimizing distribution of the mutual information is indifferent to the orientation of $\rvw_\ell$, and is characterized by a single parameter $k$, as RIQ suggests.

First,  extending \cref{eq: reconstruct w} to model quantization, where layer $\ell$ is encoded uniformly, yields
\begin{equation}
    \rvwh_\ell= \left\lceil {\rvw_\ell}/{\Delta_\ell}\right\rfloor \cdot \Delta_\ell
    \label{eq: uniform vector quant}
\end{equation}
For tractability, it is common to analyze the rate-distortion for \cref{eq: uniform vector quant} by a surrogate model in which the distortion is modeled as a random additive noise \citep{9439875, 1424312}. Yet, when considering angular deviation such representation hinders the rate-distortion analysis since the relation between additive noise to cosine distance that we wish to examine is intricate. 
Accordingly, we suggest a reparameterization to the additive noise model, which represents quantization as \emph{random rotation} (and scale) of the weights in each layer. This enables to analyze the rate-distortion of \cref{eq: uniform vector quant} for the deviation in \cref{eq: cosine dist def} in the sequel. 
\begin{model}\label{model}
Let $\rvw_\ell$ be the weights of layer $\ell$, and let $\rvwh_\ell$ denote their quantized representation. Let $\theta_\ell$ be a random rotation angle from $\rvw_\ell$ to  $\rvwh_\ell$, such that $\inp{\frac{\rvw_\ell}{\Vert \rvw_\ell \Vert }}{\frac{\rvwh_\ell}{\Vert \rvwh_\ell \Vert}} = \cos(\theta_\ell)$, and let  $\rmU(\theta_\ell \big| \rvw_\ell)\in SO\prnt{n_\ell}$ be  a random orthogonal transformation corresponding to a random  rotation that is $\theta_\ell$ away from $\rvw_\ell$. Then, 
\begin{equation}
\rvwt_\ell=  \Vert \rvwh_\ell  \Vert \cdot \rmU(\theta_\ell \big| \rvw_\ell) \frac{\rvw_\ell}{\Vert \rvw_\ell \Vert}
    \label{eq: quantization to rotation}
\end{equation}
models the quantized weights $\rvwh_\ell$.
\end{model}
Intuitively, $\rmU(\theta_\ell \big| \rvw_\ell)$ randomly rotates any given vector uniformly on a sphere, where one degree of freedom is lost due to the requirement of being $\theta_\ell$ away from $\rvw_\ell$. To obtain $\rvwh_\ell$ in \cref{eq: uniform vector quant}, the realization of $\rmU(\theta_\ell \big| \rvw_\ell)$ should rotate the unit  vector $\rvw_\ell/\Vert \rvw_\ell \Vert$ in the plane generated by $\rvw_\ell$ and $\rvwh_\ell$, and then, stretches it into the length $\Vert\rvwh_\ell\Vert$ (see \Cref{fig: surrogate model illustration} for illustration). 

In other words, this model describes a random vector $\rvwt_\ell$  that is uniformly distributed on a cone  that is $\theta_\ell$ away from $\rvw_\ell$, for which a single realization matches \cref{eq: uniform vector quant}.  
The merit of this model is its tractable analysis, from which \emph{spherically symmetric distribution} emerges to depict the quantized weights \citep[Definition 2.1]{fang2018symmetric}. 
Accordingly, each layer obtains a randomly rotated version of $\rvw_\ell$, which translates to a joint rotation at angle $\theta_{[1:L]}$ of all the parameters by Corollary~\ref{coro: distortion to parameters distortion}. Consequently, in the high rate regime, where the support of $\theta_\ell$ is sufficiently small, the distortion and the deviation decrease at same scale with $k$  by Proposition~\ref{prop: monotonically decreasing}.

\begin{proposition}
Let $\rvw_\ell$ be the weights of layer $\ell$, and let $\rvwt_\ell$ model the quantized representation of those weights, modeled by \cref{eq: quantization to rotation}. Then,  \(\rvwt_\ell \big| \rvw_\ell\) has a spherical (rotation-invariant) distribution. 
\label{prop: rotation invariance of rvwh}
\end{proposition}
\ifshort
A detailed proof is given in \Cref{sec. proof of prop1}.
\fi
Essentially, the strength of Proposition~\ref{prop: rotation invariance of rvwh} is twofold. First, it proves that the distribution of  $\rvwt_\ell \big| \rvw_\ell$ in each layer does not change when arbitrary rotations are applied to it.  Second, it holds for any distribution of $\rvw_\ell$ and $\rtheta_\ell$. This universality is substantial when considering various models and tasks. 
The following theorem extends Proposition~\ref{prop: rotation invariance of rvwh} to multiple layers, including the model activation and non-linearity, showing that spherical distribution is also the rate-distortion minimizing distribution. 
\begin{theorem}
Let $\gM$ be a NN model with $L$  layers whose weights are $\rvw_{[1:L]}$, and let $\rvwt_{[1:L]}$ be their quantized representation. Then, the unique minimizing distribution $p (\tilde{\ww}_{[1:L]} \big| \ww_{[1:L]})$ of the rate-distortion function
\begin{equation}
   R(D) = \min_{ \stackrel{p\prnt{\tilde{\ww}_{[1:L]} \big| \ww_{[1:L]}}:}{\E \brkt{ d_{f,x}(\rvw_{[1:L]} ,\rvwt_{[1:L]} )} \leq D}} I \prnt{\rvw_{[1:L]} ;\rvwt_{[1:L]} } 
    \label{eq: R(D) for NNs}
\end{equation}
is a product distribution constructed as the product of the layers’ spherical distribution. Consequently, the infimum rate is characterized by a single parameter. 
\label{theorem: prod single rotation-invariant}
\end{theorem}
\ifshort 
\input{sections/proofs/theorem1}

Remarkably, the joint minimizing distribution of the model's weights $p\prnt{\tilde{\ww}_{[1:L]} \big| \ww_{[1:L]}}$ is also spherical since any partitioning of spherical distribution (naturally occurring by the model's layers) remains spherical \citep[Theorem 2.6]{fang2018symmetric}. Further, due to the convexity of the mutual information, it is beneficial to consider as many partitions as possible, which can only reduce the mutual information. Practically, however, running the model layer-by-layer, where each layer is quantized by a single scalar, has merit due to its simplicity.   
We point out that the minimizing distribution cannot be found explicitly without any assumption on the distribution of the weights.  Interestingly, the optimal solutions presented by  \citet{gao2019rate, isik2022information} for the Gaussian and Laplace distributions coincide with our rotation-invariant observation as these distributions are spherical distributions as well. Characterizing  a universal rate-distortion problem is substantial as it reveals that it is sufficient to consider only rotation-invariant solutions for any model and task. 
\fi

%% file: sections/proofs/coro1.tex
\begin{proof}
Let $\rvw_\ell$  be the realization of the weights vector of layer $\ell$, and $\rvwh_\ell$ be the quantized representation of those weights, where $\theta_\ell$ denotes the angle between those vectors. Before diving into the cosine distance analysis, let us revisit the mean squared error analysis of the uniform quantizer in  \Cref{sec. scalar quant}, and extend it to the multivariate case. In this case, the distortion is
 \begin{align}
  \Vert \rvw_\ell- \rvwh_\ell \Vert^2 & = n_\ell \cdot \frac{1}{n_\ell} \sum_{i=1}^{n_\ell} \vrnt{ \rw_{\ell,i}- \hat{\rw}_{\ell,i} }^2  &\nonumber \\
  &\stackrel{(a)}{\to} n_\ell \cdot \E \vrnt{ \rw_{\ell,j}- \hat{\rw}_{\ell,j}  }^2  &\nonumber \\
  &\stackrel{(b)}{=}   n_\ell \cdot \Delta^2_\ell / 12  \label{eq. mse vector quant}
\end{align}
where (a) follows from the law of large numbers, and (b) follows by the analysis of the scalar uniform quantizer, given in \citep[Ch. 24.1]{polyanskiy2022information}. 

Next, let us focus on the high-rate regime, where  $R = O(\log^c(\Vert \rvw_\ell \Vert))$ for $c>1$, for which the resulting error bound is $o(\Vert \rvw_\ell \Vert)$.
For analyzing the cosine distance between $\rvw_\ell$ and $\rvwh_\ell$ in the high-rate regime, we notice that
$$\Vert \rvw_\ell - \rvwh_\ell \Vert^2 = \Vert \rvw_\ell \Vert^2 + \Vert \rvwh_\ell \Vert^2 - 2\Vert \rvw_\ell \Vert\cdot \Vert \rvwh_\ell \Vert \cos(\rtheta_\ell)$$
 Assuming   $\Vert \rvwh_\ell \Vert = \Vert \rvw_\ell \Vert + o(\Vert \rvw_\ell \Vert)$, yields
\begin{align*}
  \Vert \rvw_\ell - \rvwh_\ell \Vert^2 &= 2\Vert \rvw_\ell \Vert^2 + o(\Vert \rvw_\ell \Vert^2) - 2\Vert \rvw_\ell \Vert^2 \cos(\rtheta_\ell) + o(\Vert \rvw_\ell \Vert^2)&\\
  & =  2\Vert \rvw_\ell \Vert^2\cdot\prnt{1- \cos(\rtheta_\ell)} + o(\Vert \rvw_\ell \Vert^2).  
\end{align*}
Hence, normalizing both sides by $2\Vert \rvw_\ell \Vert^2$, we obtain that 
\begin{equation}
\prnt{1-\cos(\rtheta_\ell)} = \frac{\Vert\rvw_\ell - \rvwh_\ell \Vert^2}{2\Vert \rvw_\ell \Vert^2} + o(1).
    \label{eq: lemma 1}
\end{equation}
Combining the analysis of \cref{eq. mse vector quant} with \cref{eq: lemma 1}, we obtain
$$\prnt{1- \cos(\rtheta_\ell)} = \frac{\Vert \rvw_\ell - \rvwh_\ell \Vert^2}{2\Vert \rvw_\ell \Vert^2} + o(1) = \frac{\Delta_\ell^2 \cdot n_\ell}{24 \cdot \Vert \rvw_\ell \Vert^2} +o(1).$$
By denoting $\eps_\ell = 1- \cos(\rtheta_\ell)$, and omitting the little order $o(1)$, the lemma follows.
\end{proof}

%% file: sections/proofs/prop2.tex
\begin{proof}
The layer whose quantization error converges last to $\teps$ dictates when to stop the search. Specifically, when $k$ is sufficiently large in \cref{eq: scale}, the error in layer $\ell^*$ reaches $\sqrt{\eps_{\ell^*}} = o(\teps) + \teps$, where $o(\cdot)$ denotes little order of magnitude. That is where $\teps$ becomes dominant. At this point, we say that the error has converged for all layers (as it converged even at the largest layer $\ell^*$). Since $\teps\leq 1$ in the cosine distance criterion, we choose the little order of magnitude to be $o(\teps) = \teps\cdot\sqrt{\teps}$, and hence,  $k$ can be bounded from above by 
$$\frac{1}{k}\sqrt{n_{\ell^*}/24} + \teps \geq o(\teps) + \teps,$$ 
which happens when $k\leq \sqrt{n_{\ell^*}/24}/(\teps\cdot\sqrt{\teps})$.

In our experiments, we let $\teps=0.01$, hence, the upper limit is simply $k\leq 1000\cdot \sqrt{n_{\ell^*}/24}$.~\footnote{When the original weights are represented with $R$ bit symbols, then, choosing $\teps = 0$, yields a trivial upper bound $k_{\max}$, which is the largest number that can be represented with $R$ bits, e.g., $\round{k_{\max}} \leq 2^{31}$, when using 32 bits integer.}
  For a lower bound, we use again the fact that $\eps_\ell\leq 1$. Thus, focusing on layer $\ell^*$, we observe that
$$\frac{1}{k}\cdot \sqrt{n_{\ell^*}/24} + \teps \leq 1,$$
which happens as long as  $k \geq \sqrt{n_{\ell^*}/24}/(1-\teps)$. This completes the proof.
\end{proof}

As mentioned, the power of  $\eps_0$ is substantial. On the one hand, $\eps_0$ shifts the bin-width $\Delta_\ell$ from the exact rotation-invariant solution. On the other hand, without this term, the quantization rate might get too high, and hence, it has practical merit. To examine its influence, we used RIQ on ResNet50 with different values of $\eps_0$, recording the deviation as a function of the $k$ parameter and the achieved compression (see \Cref{fig: eps_0 influence}). Interestingly, large values of $\eps_0$ did not reach the target deviation (which is 0.005 in our experiments). Yet, since enforces lower rates, then, it attains a good compression ratio. On the other hand, picking small $\eps_0$ is beneficial in terms of model deviation, which is crucial for accuracy. The best value tradeoffs between these options. That is, picking the largest value of $\eps_0$ that satisfies the deviation requirement. 

%% file: sections/proofs/prop1.tex
\begin{proof}
By \citet[Theorem 4.3]{fang2018symmetric}, a necessary and sufficient condition for  $\rvwt_\ell\big| \rvw_\ell$ to have a spherical rotation-invariant distribution on a cone  that is $\theta_\ell$ away from $\rvw_\ell$ is when
$$p\prnt{\left.\inp{\rvwt_\ell}{ \rvv_1}\right| \inp{\rvwt_\ell}{ \rvv_2}, \rvw_\ell} \stackrel{d}{=} p\prnt{\left.-\inp{\rvwt_\ell}{ \rvv_1}\right| \inp{\rvwt_\ell}{\rvv_2}, \rvw_\ell},$$ 
for any pair of perpendicular vectors $\rvv_1 \neq 0$ and $\rvv_2 \neq 0$ that are orthogonal to $\rvw_\ell$. 

Consider the model in \cref{eq: quantization to rotation},  any orthogonal transformation $\rmU(\rtheta_\ell \big| \rvw_\ell)$ can be represented by an orthonormal basis, obtained by the  Gram-Schmidt process. That is, finding two orthonormal vectors $\rvu_1$ and $\rvu_2$ that span the plane of rotation generated by $\rvw_\ell$ and some $\rvw^\prime_\ell$ that is $\rtheta_\ell$ away from $\rvw_\ell$, and then, extend this basis to $\R^{n_\ell}$. This allows us to consider the rotation  in the plane generated by those vectors, with respect to the extended basis \cite{598782}. Accordingly,  let $\rvu_1 = \frac{\rvw_\ell}{\Vert \rvw_\ell \Vert}$ and $\rvu_2 = \frac{\rvw^\prime_\ell - \langle \rvu_1, \rvw^\prime_\ell \rangle \rvu_1 }{\Vert \rvw^\prime_\ell - \langle \rvu_1, \rvw^\prime_\ell \rangle \rvu_1  \Vert}$, then 
\begin{equation}
    \rmU(\rtheta_\ell \big| \rvw_\ell) = \rmI_{n_\ell} - \rvu_1\rvu_1^{\etT} - \rvu_2 \rvu_2^{\etT} + \brkt{\rvu_1 , \rvu_2} \rmR_{\rtheta_\ell} \brkt{\rvu_1 , \rvu_2}^\etT,
    \label{eq: rotation u}
\end{equation}
where $\rmI_{n_\ell}$ is the $n_\ell\times n_\ell$ identity matrix and $\rmR_{\rtheta_\ell}$ is the rotation matrix
\[
 \rmR_{\rtheta_\ell} =
  \left[ {\begin{array}{cc}
    \cos(\rtheta_\ell) & -\sin(\rtheta_\ell) \\
    \sin(\rtheta_\ell) & \cos(\rtheta_\ell) \\
  \end{array} } \right],
\]
that rotates at a scalar angle $\rtheta_\ell$, and $\brkt{\rvu_1 , \rvu_2}$ is $n_\ell \times 2$ matrix whose columns are $\rvu_1$ and $\rvu_2$, respectively.  Plugging \cref{eq: rotation u} to \cref{eq: quantization to rotation}, and noting that $\frac{\rvw_\ell}{\Vert \rvw_\ell \Vert} = \rvu_1$, we obtain
 \begin{equation}
     \rvwt_\ell = \Vert \rvwh_\ell \Vert \cdot \prnt{\cos(\rtheta_\ell) \rvu_1 + \sin(\rtheta_\ell)\rvu_2} 
     \label{eq: rvwh ell}
 \end{equation}
To simplify notation, let  $\rvu \triangleq \prnt{\cos(\rtheta_\ell) \rvu_1 + \sin(\rtheta_\ell)\rvu_2}$, and note that for any perpendicular pair $\rvv_1, \rvv_2$ that are orthogonal to $\rvw_\ell$, the vector $\rvu$ can be decomposed to $\rvu = \rvu^\parallel + \rvu^\perp$, where $\rvu^\parallel$ resides in the plane generated by $\rvv_1$ and $\rvv_2$, and $\rvu^\perp$ resides in the null-space of this plane. For illustration, see \Cref{fig: projection illustration}. Hence, 

\begin{figure*}[]
\centering
 \subfigure[]{
 \centering
    \includegraphics[width=0.3\textwidth]{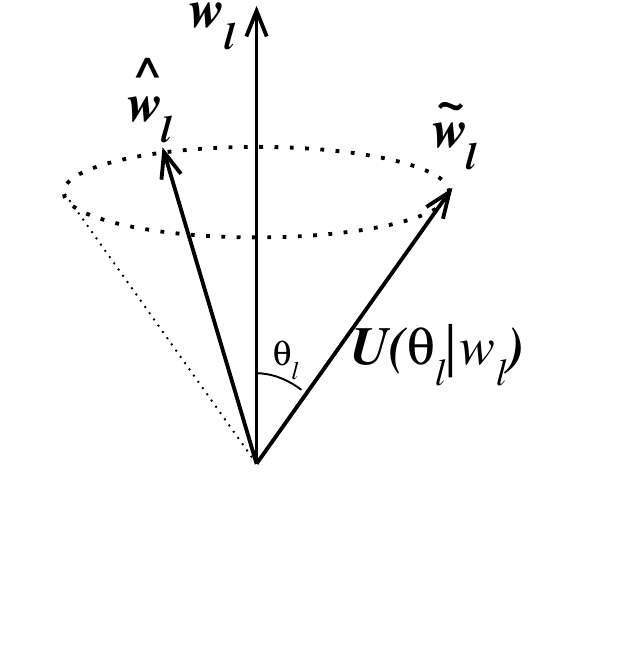}
    \label{fig: surrogate model illustration}
 }
 \hfill
 \subfigure[]{
 \centering
    \includegraphics[width=0.3\columnwidth]{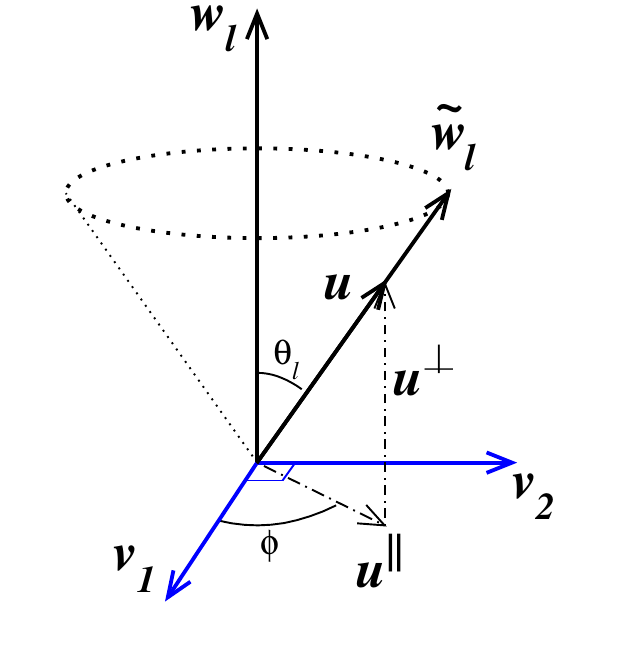}
    \label{fig: projection illustration}
 } 
 \caption{(a) Illustration of the surrogate model in \Cref{model}. Is this illustration, the quantized weights are modeled by the result of an orthogonal transformation $\rmU(\theta_\ell | \rvw_\ell)$ which rotates the vector $\rvw_\ell$ randomly onto a ring that is $\theta_\ell$ away from $\rvw_\ell$. Note that the true quantization results also lies in this ring. (b) Illustration of the projection of $\rvwt_\ell$ onto the arbitrary perpendicular vectors $\rvv_1$ and $\rvv_2$.}
\end{figure*}

\begin{equation}
     \rvwt_\ell = \Vert \rvwh_\ell \Vert \cdot (\rvu^\parallel + \rvu^\perp) 
     \label{eq: rvwh ell simple}
 \end{equation}

Accordingly, we have
\begin{align}
    & p\prnt{\left.\inp{\rvwt_\ell}{ \rvv_1}\right| \inp{\rvwt_\ell}{ \rvv_2}, \rvw_\ell} &\\
    & \stackrel{(a)}{=} p\prnt{\left.\inp{\Vert \rvwh_\ell \Vert \cdot (\rvu^\parallel + \rvu^\perp)}{ \rvv_1} \right| \inp{\Vert \rvwh_\ell \Vert \cdot (\rvu^\parallel + \rvu^\perp)}{\rvv_2 }, \rvw_\ell }&\nonumber\\
    & \stackrel{(b)}{=} p\prnt{\left.\inp{\Vert \rvwh_\ell \Vert \cdot \rvu^\parallel}{ \rvv_1} \right| \inp{\Vert \rvwh_\ell \Vert \cdot \rvu^\parallel}{\rvv_2 }, \rvw_\ell}&\nonumber\\
   &\stackrel{(c)}{=} p\prnt{\left.\Vert \rvwh_\ell \Vert \cdot \Vert \rvu^\parallel \Vert \cdot  \cos(\phi) \right| \Vert \rvwh_\ell \Vert \cdot \Vert \rvu^\parallel \Vert  \cdot \sin(\phi), \rvw_\ell}&\nonumber\\
   &\stackrel{(d)}{=} p\prnt{\left.\Vert \rvwh_\ell \Vert \cdot \Vert \rvu^\parallel \Vert \cdot \cos(\pi - \phi) \right| \Vert \rvwh_\ell \Vert \cdot \Vert \rvu^\parallel \Vert \cdot \sin(\pi - \phi), \rvw_\ell}&\nonumber\\
   &\stackrel{(e)}{=} p\prnt{\left. - \Vert \rvwh_\ell \Vert \cdot \Vert \rvu^\parallel \Vert \cdot \cos(\phi) \right| \Vert \rvwh_\ell \Vert \cdot \Vert \rvu^\parallel \Vert \cdot \sin(\phi), \rvw_\ell}&\nonumber\\
   & = p\prnt{-\left.\inp{\rvwh_\ell}{ \rvv_1}\right| \inp{\rvwh_\ell}{ \rvv_2}, \rvw_\ell}
\end{align}
where (a) follows by \cref{eq: rvwh ell simple}. (b) follows by the linearity of the inner product and since $\rvu^\perp$ is perpendicular to both $\rvv_1$ and $\rvv_2$ (c) follows due to the orthogonality of the basis $\rvv_1$ and $\rvv_2$, where $\phi$ is the angle between $\rvu^\parallel$ and $\rvv_1$. (d) follows since the angle between $\rvu^\parallel$ and an arbitrary $\rvv_1$ is arbitrary, and hence, every angle has the same distribution. (e) follows by  trigonometric identities for the cosine and sine function. Thus, Proposition~\ref{prop: rotation invariance of rvwh} follows. 
\end{proof}

%% file: sections/proofs/theorem1.tex
\begin{proof}

Consider the rate-distortion function in \cref{eq: R(D) for NNs}. 
Since  $\rvw_{[1:\ell-1]} \to \rvw_\ell \to  \tilde{\rvw}_\ell$ form a Markov chain (i.e.,  given $\rvw_\ell$ then $\tilde{\rvw}_\ell$  and $\rvw_{[1:\ell-1]}$ are independent.), then,  by the properties of the mutual information, we have
\begin{align}
    I &\prnt{\rvw_{[1:L]} ;\rvwt_{[1:L]} } = H\prnt{\rvw_{[1:L]}} -  H\prnt{\left. \rvw_{[1:L]} \right| \rvwt_{[1:L]}}&\\
    & = \sum_{\ell = 1}^L   H\prnt{\left.\rvw_\ell\right|\rvw_{[1:\ell-1]}} - \sum_{\ell = 1}^L H\prnt{\left. \rvw_\ell \right| \rvwt_{[1:L]}, \rvw_{[1:\ell-1]} } &\\
    & \geq \sum_{\ell = 1}^L   H\prnt{\left.\rvw_\ell\right|\rvw_{[1:\ell-1]}} -  \sum_{\ell = 1}^L H\prnt{\left. \rvw_\ell \right| \rvwt_\ell, \rvw_{[1:\ell-1]} } & \label{eq: R(D) conditoning} \\
    & = \sum_{\ell = 1}^L I \prnt{\left.\rvw_\ell ;\rvwt_\ell \right| \rvw_{[1:\ell-1]} } & \\
    & = \sum_{\ell = 1}^L I \prnt{\rvw_\ell ;\rvwt_\ell} - \sum_{\ell = 1}^L I \prnt{\rvwt_\ell ; \rvw_{[1:\ell-1]}}  \label{eq: makov chain independence}&\\
    & \geq \sum_{\ell = 1}^L I \prnt{\rvw_\ell ;\rvwt_\ell} - \sum_{\ell = 1}^L H\prnt{\rvw_{[1:\ell-1]}} \label{eq: mutual information sum}&\\
    & \geq \sum_{\ell = 1}^L R(D_\ell) -  \sum_{\ell = 1}^L H\prnt{\rvw_{[1:\ell-1]}} \label{eq: rate distortion sum} &
\end{align}
where \cref{eq: R(D) conditoning} follows since conditioning reduces entropy. Note, however, that \cref{eq: R(D) conditoning} can be attained with equality by letting $p\prnt{\left. \rvw_{[1:L]} \right| \rvwt_{[1:L]} } = \prod_{\ell=1}^L p\prnt{ \rvw_\ell \big|\rvwt_\ell}$. Consequently,  the minimizing distribution in \cref{eq: R(D) for NNs} is a product distribution \citep[Theorem 6.1 (2)]{polyanskiy2022information}. \Cref{eq: makov chain independence} follows since $\rvwt_\ell$  and $\rvw_{[1:\ell-1]}$ are independent given $\rvw_\ell$. 

Since $H\prnt{\rvw_{[1:\ell-1]}}$ is a function of $p\prnt{\rvw_{[1:\ell-1]}}$, which we cannot optimize, our focus is on optimizing $I \prnt{\rvw_\ell ;\rvwt_\ell}$. This  requires formulating the relation between the deviations  $D$ and $D_\ell$ of each layer $\ell$, and hence, the resulting rate $R(D_\ell)$ for each layer. Moreover, since each layer obtains a different rate, it implies that the optimal solution is indeed a \emph{mixed-precision solution}, where each layer can be considered independently, and hence,  the minimizing distribution is product distribution.    


Consider the model deviation. Let $p\prnt{\rvwt_1 \big| \rvw_1}\cdot p\prnt{\rvwt_2 \big| \rvw_2} \cdots  p\prnt{\rvwt_L \big| \rvw_L}$ be a distribution that satisfies the deviation requirement $D$, 
for which the induced cosine distance (distortion) in each layer $\ell$ is at most $\eps_\ell$, for $\eps_\ell = 1 - \cos(\theta_\ell)$. 
By Corollary~\ref{coro: distortion to parameters distortion}, the distortion over the entire parameters is hence $\eps_{[1:L]} \triangleq 1- \cos(\theta_{[1:L]}) = \sum_{\ell=1}^L \frac{\Vert \rvw_\ell \Vert^2}{\Vert \rvw_{[1:L]} \Vert^2} \eps_\ell$. Note that $\eps_{[1:L]} $ is a convex combination of $\eps_{\ell}$ as $\sum_{\ell=1}^L \frac{\Vert \rvw_\ell \Vert^2}{\Vert \rvw_{[1:L]} \Vert^2} = 1$.
Assuming $\eps_{[1:L]} \leq D$,  then due to the convexity of the cosine distance for $\vert\rtheta_\ell\vert \leq \pi/2 $, by Jensen inequality  $\prnt{1-\cos\prnt{\sum_{\ell=1}^L \frac{\Vert \rvw_\ell \Vert^2}{\Vert \rvw_{[1:L]} \Vert^2} \rtheta_\ell}} \leq \sum_{\ell=1}^L \frac{\Vert \rvw_\ell \Vert^2}{\Vert \rvw_{[1:L]} \Vert^2}\prnt{1-\cos(\rtheta_\ell)}$.  
In words,  convex combination of the angles also satisfies the deviation constraint $D$.

Next, to address $p(\rvwt_\ell | \rvw_\ell)$, 
let us represent the transformation $\rmU(\rtheta_\ell \big| \rvw_\ell)$ by a pair of orthonormal vectors $\rvu_1$ and $\rvu_2$, using  Gram-Schmidt process (see  \Cref{sec. proof of prop1} for details). These vectors span the plane of rotation generated by $\rvw_\ell$ and some $\rvw^\prime_\ell$ that is $\rtheta_\ell$ away from $\rvw_\ell$. This simplifies \cref{eq: quantization to rotation} to 
\begin{equation}
  \rvwt_\ell = \Vert \rvwh_\ell \Vert \cdot \prnt{\cos(\rtheta_\ell) \rvu_1 + \sin(\rtheta_\ell)\rvu_2}.  
\end{equation}
Thus, by Proposition~\ref{prop: rotation invariance of rvwh}, given $\rvw_\ell$ (and hence, $\rvu_1$), the probability of $\rvwt_\ell$ is determined by the probability of the rotation angle $\rtheta_\ell$ and the norm $\Vert \rvwh_\ell  \Vert$. Specifically, for any vector $\rvs_\ell\in \R^{n_\ell}$, the density function of this product, if exists, is \citep[Ch. 4.1]{springer_1979}
\begin{multline}
    p_{\rvwt_\ell |\rvw_\ell} (\rvs_\ell) = \int_{0}^\infty p_{\Vert \rvwh_\ell \Vert |\rvw_\ell}\prnt{h} \\
	\cdot p_{\prnt{\cos(\rtheta_\ell) \rvu_1 + \sin(\rtheta_\ell)\rvu_2} |\rvw_\ell}\prnt{\rvs_\ell/h}\cdot\frac{1}{h}\mathrm{d}h,
    \label{eq: prod distribution}
\end{multline}
where the rotation  $\rtheta_\ell$ occurs on $\R^2$, rotating about $(n_\ell -2)$-dimensional subspace. Further, note that the dimension $n_\ell$ is dictated only by the given $\rvw_\ell$. Apparently, since each layer $\ell$ resides at a different dimension $n_\ell$, it is impossible to consider the convex combination of the layers' distribution directly, as done for the vector case,  e.g., as considered in \citet[Ch. 5]{polyanskiy2014lecture}. Nevertheless, since the rotation of $\rtheta_\ell$ is done on $\R^2$ in each layer $\ell$, 
it is still beneficial to consider a convex combination of $\rtheta_\ell$ distributions over the layers, to allow a similar treatment to \citet[Ch. 5]{polyanskiy2014lecture}, as follows.

To bound the mutual information, the density of $\prnt{\cos(\rtheta_\ell) \rvu_1 + \sin(\rtheta_\ell)\rvu_2} |\rvw_\ell$ should be expressed in terms of the density of $\cos(\rtheta_\ell) | \rvw_\ell$. 
Accordingly, by the transformation of random variables formula we have,
\begin{equation}
    p_{\prnt{\cos(\rtheta_\ell) \rvu_1 + \sin(\rtheta_\ell)\rvu_2} |\rvw_\ell}\prnt{\rvs_\ell/h} = p_{\cos\prnt{\rtheta_\ell} |\rvw_\ell}\prnt{\rvu_1^\etT \rvs_\ell /h}.
    \label{eq: trans to theta}
\end{equation}
Considering the high rate regime, where $\Delta_\ell$ is sufficiently small, and thus, $\Vert \rvwh_\ell \Vert = \Vert \rvw_\ell \Vert + o(1)$, then, the density function in \cref{eq: prod distribution} becomes
\begin{align*}
\small
  &p_{\rvwt_\ell \big| \rvw_\ell} (\rvs_\ell) = p_{\left. \Vert \rvwh_\ell \Vert \cdot \prnt{\cos(\rtheta_\ell) \rvu_1 + \sin(\rtheta_\ell)\rvu_2} \right| \rvw_\ell} (\rvs_\ell)  &\\
  &\stackrel{}{=} \int_{0}^\infty p_{\Vert \rvwh_\ell \Vert |\rvw_\ell}\prnt{h} \cdot p_{\prnt{\cos(\rtheta_\ell) \rvu_1 + \sin(\rtheta_\ell)\rvu_2} |\rvw_\ell}\prnt{\rvs_\ell/h}\cdot\frac{1}{h}\mathrm{d}h &\\
  &\stackrel{(a)}{=} \int_{0}^\infty p_{\Vert \rvwh_\ell \Vert |\Vert\rvw_\ell\Vert}\prnt{h} \cdot p_{\prnt{\cos(\rtheta_\ell) \rvu_1 + \sin(\rtheta_\ell)\rvu_2} |\rvw_\ell}\prnt{\rvs_\ell/h}\cdot\frac{1}{h}\mathrm{d}h &\\
   &\stackrel{(b)}{\approx}\int_{0}^\infty \delta\prnt{ h - \Vert\rvw_\ell\Vert } \cdot p_{\prnt{\cos(\rtheta_\ell) \rvu_1 + \sin(\rtheta_\ell)\rvu_2} |\rvw_\ell}\prnt{\rvs_\ell/h}\cdot\frac{1}{h}\mathrm{d}h &\\
  &\stackrel{(c)}{=} \int_{0}^\infty \delta\prnt{ h - \Vert\rvw_\ell\Vert} \cdot p_{\cos\prnt{\rtheta_\ell} |\rvw_\ell}\prnt{\rvu_1^\etT \rvs_\ell /h}\cdot\frac{1}{h}\mathrm{d}h &\\
  &\stackrel{(d)}{=}   p_{\cos\prnt{\rtheta_\ell} |\rvw_\ell}\prnt{\rvu_1^\etT \rvs_\ell /\Vert\rvw_\ell\Vert}\cdot\Vert\rvw_\ell\Vert^{-1}  &
\end{align*}
where (a)  follows since the norm $\Vert \rvw_\ell\Vert$ is a function of the given $\rvw_\ell$.  
(b) follows since the uncertainty about $\Vert \rvwh_\ell \Vert$ given $\Vert \rvw_\ell \Vert$ is negligible, and hence, $p_{\Vert \rvwh_\ell \Vert | \Vert \rvw_\ell \Vert}(h) \approx \delta(h - \Vert \rvw_\ell\Vert)$, i.e., the conditional density is approximately the Dirac delta function.  (c) follows by \cref{eq: trans to theta}. (d) follows by the characteristics of the Dirac delta function. 

Hence, it is possible to consider a convex combination of $p\prnt{\rvwt_\ell \big| \rvw_\ell}$ over the layers, where each layer is embedded in possibly different $n_\ell$,  by considering a convex combination of the rotations' probability $p_{\cos\prnt{\rtheta_\ell} |\rvw_\ell}$, since all rotations are done in $\R^2$. Accordingly, let 
\begin{multline}
    \bar{p}_{\rvwt_{[1:L]} \big| \rvw_{[1:L]}} (\rvs_{[1:L]}) \triangleq \sum_{\ell=1}^L \frac{\Vert \rvw_\ell \Vert^2}{\Vert \rvw_{[1:L]} \Vert^2} \\
	\quad\cdot p_{\cos\prnt{\rtheta_\ell} |\rvw_\ell}\prnt{\frac{\rvw_\ell^\etT \rvs_\ell }{\Vert\rvw_\ell\Vert^2}}\cdot\Vert\rvw_\ell\Vert^{-1}.
\label{eq: p bar}
\end{multline}
Then, by the convexity of the rate-distortion function \citet[Theorem 2.7.4]{thomas2006elements}, $\bar{p}\prnt{\rvwt_{[1:L]} \big| \rvw_{[1:L]}}$ can only reduce the mutual information in \cref{eq: mutual information sum}. Specifically, 
$$\sum_{\ell = 1}^L I_{p\prnt{\rvwt_\ell \big| \rvw_\ell}} \prnt{\rvw_\ell ;\rvwt_\ell } \geq L \cdot I_{\bar{p}\prnt{\rvwt_{[1:L]} \big| \rvw_{[1:L]}}} \prnt{\rvw_{[1:L]} ;\rvwt_{[1:L]} }$$
where $I_p(\cdot;\cdot)$ denotes explicitly the mutual information under probability $p$. Thus, the \emph{infimum rate has a form of a scalar (single-letter) rate}.

Moreover, since averaging over more rotations should further reduce the mutual information by its convexity, then,  the minimizing $p\prnt{\rvwt_{[1:L]} \big| \rvw_{[1:L]}}$ can be chosen to be rotation-invariant \citep[Ch. 6.2]{polyanskiy2022information}. Consequently, the unique minimizing distribution $p\prnt{\rvwt_{[1:L]}}$ is also rotation-invariant. Remarkably,  \citet[Theorem 2.6]{fang2018symmetric} states that when partitioning a  spherical rotation-invariant distribution  (naturally, according to the layers $\rvwt_\ell$), then its components also have a spherical rotation-invariant distribution. This coincides with Proposition~\ref{prop: rotation invariance of rvwh}, which proves that the partitioning satisfies this property.  

Accordingly, the unique minimizing distribution $p (\rvwt_{[1:L]} \big| \rvw_{[1:L]})$ of the rate-distortion function  is a product distribution over the layers, where each term $\ell$  is a spherical rotation-invariant distribution. 

\end{proof} 

%% file: sections/05_results.tex
\begin{figure*}
\centering
 \subfigure{
 \centering
    \includegraphics[width=0.45\linewidth]{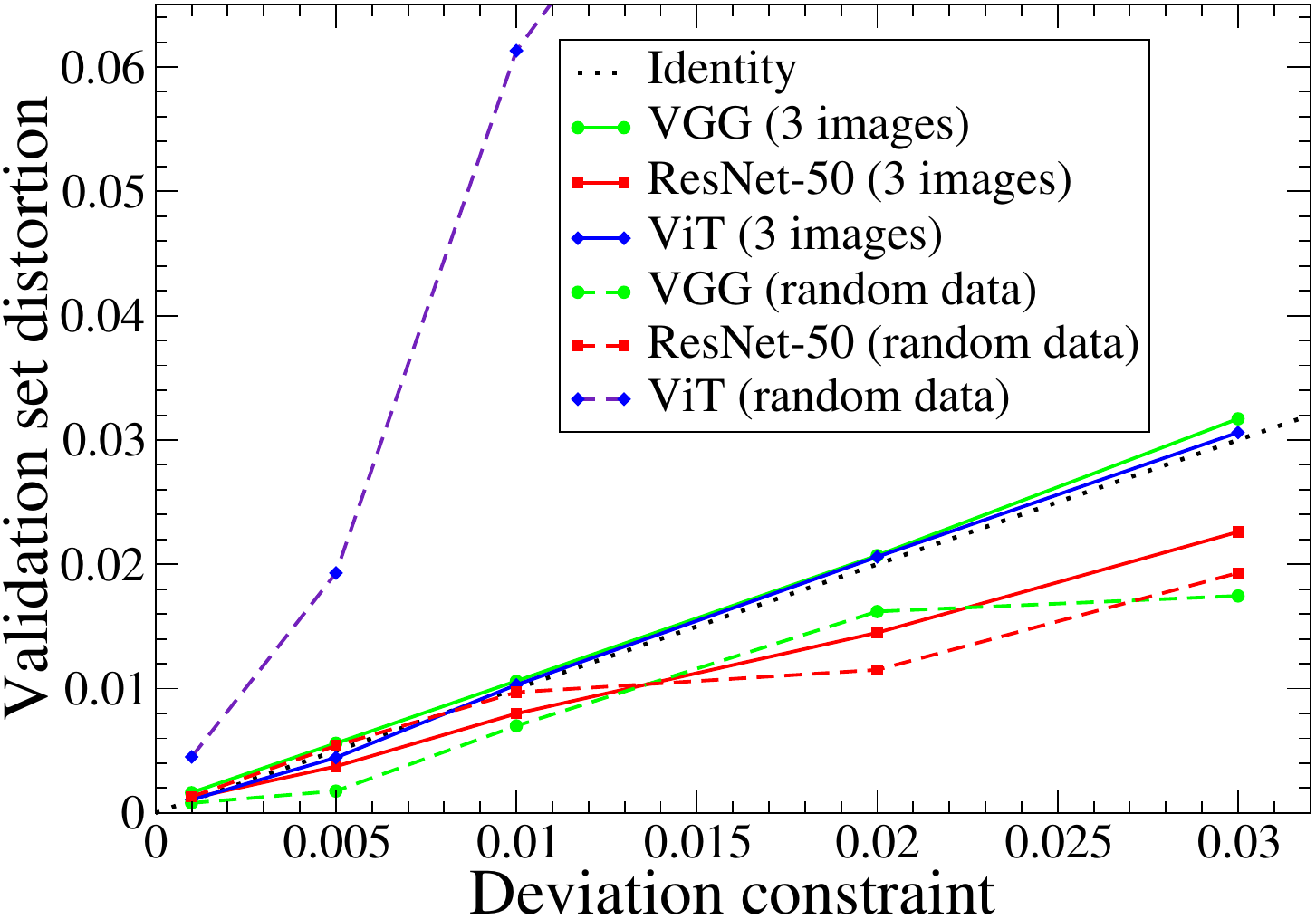}
    \label{fig:Cosine2Distortion}
 }
 \hfill
 \subfigure{
 \centering
    \includegraphics[width=0.43\linewidth]{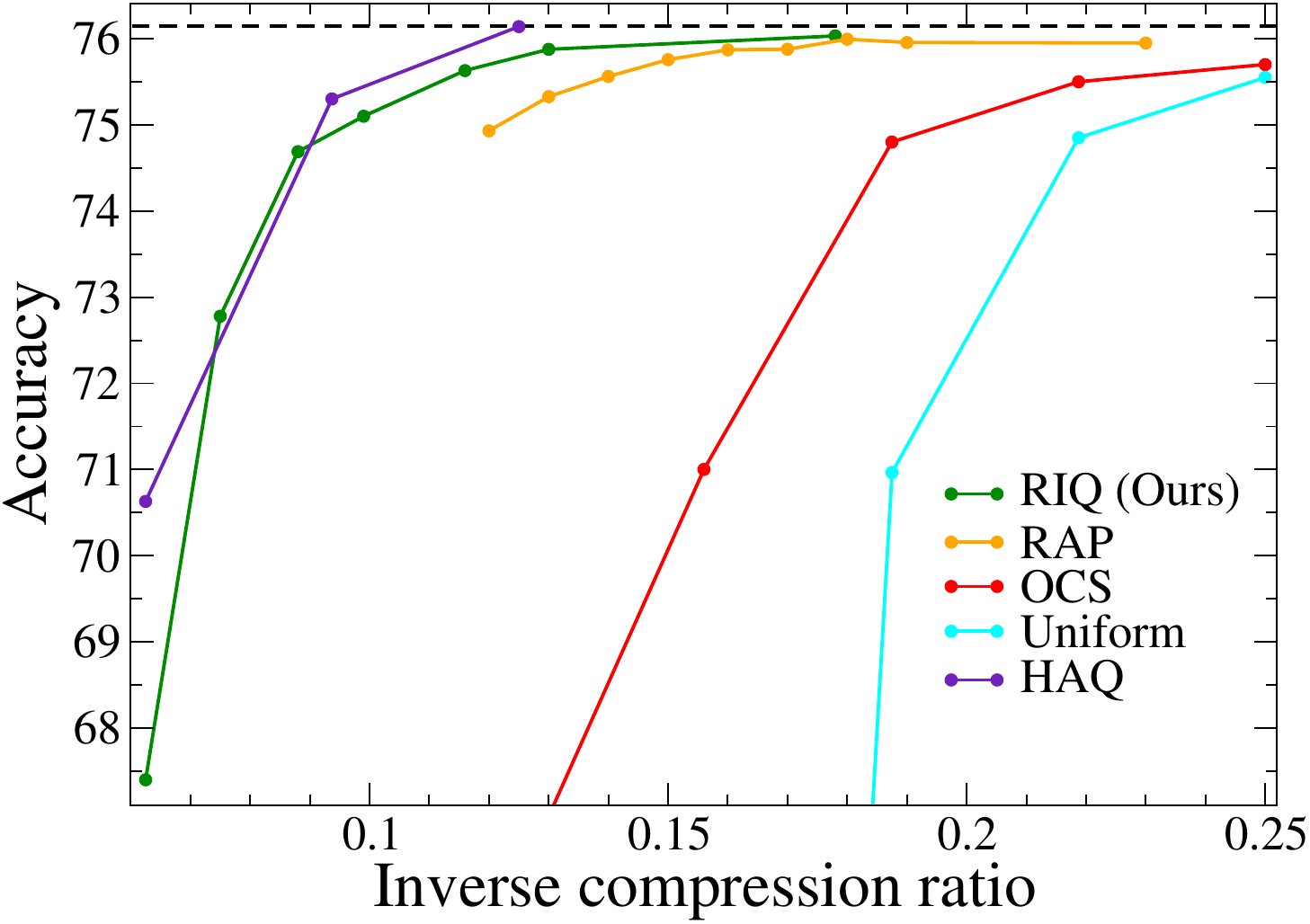}
    \label{fig:resnet50}
 }
 \caption{(a) Cosine distance of the validation dataset as a function of the deviation constraint (on calibration set). Models include VGG (green), ResNet-50 (red), and ViT (blue). (b) Accuracy vs Compression for ResNet-50 model. RIQ (green) vs. RAP (orange), OCS (red), linear quantization (cyan), and HAQ (purple). HAQ, however, requires training.}
\end{figure*}

\section{Empirical results}
In this section, we evaluate the compression ratio and model accuracy of RIQ with ANS and compare them to relevant baseline results. In all experiments, we use pre-trained models for their relevant tasks. Specifically, for classification tasks we use VGG, \citet{simonyan2014very}, ResNet-50, \citet{he2016deep}, and ViT, \citet{dosovitskiy2020image} models, from the torchvision library\footnote{\href{https://pytorch.org/vision/stable/models.html}{\nolinkurl{pytorch.org/vision/stable/models.html}}} on the ImageNet data-set (I1k). For detection tasks, we use YOLOv5, \citet{glenn_jocher_2022_6222936}, from Neural-Magic library~\footnote{\label{NM}\href{https://sparsezoo.neuralmagic.com}{\nolinkurl{sparsezoo.neuralmagic.com}}} on the COCO dataset \citep{10.1007/978-3-319-10602-1_48}. For NLP task, we use the DistilBERT model, \citet{sanh2019distilbert}, on SQuAD  dataset, \citep{rajpurkar2016squad}. Following RIQ,  an ANS encoder encodes each quantized layer to its entropy limit. An efficient implementation of ANS on GPU was demonstrated by \citet{weissenberger2019massively}, reaching impressive decoding rates. Consequently, RIQ facilitates optimizing and compressing models in a few minutes. For reproduction purposes, we provide a Python code of our algorithm as a supplementary material which includes both the quantization phase (RIQ) and compression phase (ANS)\footnote{\href{https://anonymous.4open.science/r/RIQ-6BE0/README.md}{\nolinkurl{anonymous.4open.science/r/RIQ-6BE0}}}. Additional results are given in \Cref{sec. additional results}. 

To measure the resulting deviation by \cref{eq: cosine dist def} on a validation set as a function of a deviation requirement $D$, we use two types of calibration data: (a) three real images, sampled from the training, and (b) randomly generated data that follows the Gaussian distribution. In \Cref{fig:Cosine2Distortion} we present the deviation measurements on three models: ResNet-50, VGG, and ViT, where the identity line (black-dotted) is given for reference. As we see, even a small calibration set of three images (solid lines) is sufficient to predict the deviation on the validation set.  Further, we see that the randomly generated data may not predict well the resulting deviation on the validation set, leading to either a less compressed model (ResNet-50 and VGG) or a higher deviation (ViT). 

In \Cref{fig:resnet50}, we evaluate the effect of RIQ on the accuracy and the inverse compression ratio (i.e., the reciprocal of \cref{eq:cmp_ratio}) for a pre-trained ResNet-50 model. Interestingly, the rate-distortion curve reflects well the accuracy-compression tradeoff. For comparison, we depict the accuracy-compression results of the \emph{Relaxed Advanced Pipeline} (RAP) method, \citet{pmlr-v139-hubara21a}, \emph{Outlier Channel Splitting} (OCS), \citet{zhao2019improving}, and the \emph{Hardware-aware Automated Quantization} (HAQ), \citet{wang2019haq}, which requires further training for fine-tuning. Indeed, RIQ surpasses other post-training quantization methods, yet, falls short compared to HAQ. This is since retraining the quantized model yields a different rate-distortion curve, which is out of the scope of this paper. 

\begin{table*}[ht]
  \centering 
  \scriptsize
  \begin{threeparttable}
    \caption{Comparison of Top-1 accuracy on ImageNet for various baselines.}
  \label{table:comparison}
    \begin{tabular}{c|c|l c c c }
    \specialrule{.2em}{.1em}{.1em} 
    Model & {\small Compression} & Method & Acc. (\%) & Ref. (\%) & Drop (\%)\\
    \specialrule{.2em}{.1em}{.1em}
    \multirow{8}{*}{VGG-16} 
    & \multirow{2}{*}{$\times 10.6$ (3 bits)} & GPFQ \cite{zhang2022post} \kern-1em & 70.24 & 71.59&  1.35  \\
    & & \textbf{RIQ (Ours)} & \textbf{71.58} & \textbf{71.59} & \textbf{0.01} \\

    \cline{2-6}
     & \multirow{4}{*}{$\times 8$ (4 bits)} & MSE \cite{banner2019post} \kern-1em & 70.50 &  71.60 & 1.10  \\
     & & OMSE \cite{choukroun2019low} \kern-1em & 71.48 & 73.48 & 2.00 \\
    & & GPFQ \cite{zhang2022post} \kern-1em & 70.90 & 71.59 &  0.69  \\
    & & \textbf{RIQ (Ours)} & \textbf{71.55} & \textbf{71.59} & \textbf{0.04} \\
     \cline{2-6}
     & \multirow{2}{*}{$\times 6.4$ (5 bits)} & GPFQ \cite{zhang2022post} \kern-1em & 71.05 & 71.59 &  0.54  \\
     & & \textbf{RIQ (Ours)} & \textbf{71.58} & \textbf{71.59} & \textbf{0.01} \\

    \specialrule{.2em}{.1em}{.1em} 
    \multirow{14}{*}{ResNet-50} 
     & \multirow{3}{*}{$\times 16$ (2 bits)} & SuRP with tuning \cite{isik2022information} \kern-1em  & \textbf{76.4} & \textbf{76.6} & \textbf{0.2}   \\
     & & \textbf{RIQ (Ours)} &  69.47 & 76.14 &  6.67 \\
     \cline{2-6}
    & \multirow{2}{*}{$\times 10.6$ (3 bits)} & GPFQ \cite{zhang2022post} \kern-1em  & 70.63 & 76.13 & 5.50   \\
     & & \textbf{RIQ (Ours)} &  \textbf{74.76} & \textbf{76.14} &  \textbf{1.38} \\
     \cline{2-6}
     & \multirow{7}{*}{$\times 8$ (4 bits)}  & MSE \cite{banner2019post} \kern-1em &
     73.80 & 76.10 & 2.30 \\
     & & OMSE \cite{choukroun2019low} \kern-1em &
     73.39 & 76.01 & 2.62 \\
     & & AdaRound \cite{nagel2020up} \kern-1em & 75.23 & 76.07 & 0.84 \\
     & & S-AdaQuant \cite{pmlr-v139-hubara21a} \kern-1em & 75.10 & 77.20 & 2.10 \\
     & & BRECQ \cite{li2021brecq} \kern-1em & 76.29 &  77.00 & 0.71  \\
    & & GPFQ \cite{zhang2022post} \kern-1em  & 74.35 & 76.13 & 1.78   \\
     & & \textbf{RIQ (Ours)} &  \textbf{75.61} & \textbf{76.14} &  \textbf{0.53} \\
     \cline{2-6}
     & \multirow{2}{*}{$\times 6.4$ (5 bits)} & GPFQ \cite{zhang2022post} \kern-1em & 75.26 & 76.13 & 0.87  \\
     & & \textbf{RIQ (Ours)} &  \textbf{75.95} & \textbf{76.14} &  \textbf{0.19} \\

    \specialrule{.2em}{.1em}{.1em} 
    \end{tabular}
  \end{threeparttable}
  \normalsize
\end{table*}

In \Cref{table:comparison}, we compare RIQ to relevant baseline methods on the VGG-16 and ResNet-50 models. 
In this table, we modified RIQ in Algorithm~\ref{alg: rotation inv} to minimize the deviation (accuracy drop) for a given rate requirement. 
Noticeably, RIQ outperforms most of the baselines, as it reaches the entropy limit by the ANS. Applying ANS to other baselines, however, can only degrade their compression since they consider per-channel quantization, for which the encoding table overhead becomes significant. Further, applying ANS to their layers still degrades compression as the number of unique symbols per-layer is much larger than per-channel. Moreover, some baselines quantized both weights and activation to further accelerate the models. The contribution of ANS to RIQ is discussed in \Cref{sec. additional results}. 

Typical compression ratio and score achieved by RIQ are presented in \Cref{table:typical} for a variety of models and tasks. Note that the deviation does not translate immediately to the drop in each score, as the latter is a task-specific measure. Yet, in general, the scores improve as the deviation decreases. 
To further assess the potential of RIQ, we evaluate our method on sparse models taken from the Neural-Magic\footref{NM}. Notably, the compression ratio of sparse models is significantly higher. This coincides with the conclusion that pruning is substantial for good compression \citep{isik2022information}.
\vspace{-\baselineskip}
    

\begin{table*}[ht]
\scriptsize
  \centering 
  \caption{Compression and accuracy achieved by RIQ. Models denoted by asterisk ($^\ast$) were pruned during training, before quantization. 
  }
  \label{table:typical}
  \begin{tabular}{ccccccc}
  \specialrule{.2em}{.1em}{.1em} 
  Model / Dataset & Metric & Deviation & Compression & Quant.\ Score  & Ref.\ Score &  Drop \\
  \specialrule{.2em}{.1em}{.1em} 
    VGG  / I1k & Top-1 Acc (\%) & 0.5\% & $\mathbf{\times19.4}$ & 71.3 & 71.59  & \textbf{0.29}\\
  ResNet-50 / I1k & Top-1 Acc (\%) & 0.5\% &  $\mathbf{\times 7.31}$ & 75.88 & 76.14 & \textbf{0.26}\\
  ViT / I1k  & Top-1 Acc (\%) & 0.5\% & $\mathbf{\times 6.98}$  & 81.0 & 81.07 & \textbf{0.07}\\
  YOLO / COCO  & mAP@.5 & 0.3\% & $\mathbf{\times 8.34}$ & 54.7 & 55.7 & \textbf{1.0}\\
  DistilBERT / SQuAD & F1 & 0.025\% & $\mathbf{\times 7.96}$ & 85.0 & 85.08 & \textbf{0.08} \\
  \hline
  VGG$^\ast$(75\%) / I1k & Top-1 Acc (\%) & 0.5\% & $\mathbf{\times 52.9}$ & 69.34 & 69.73 & \textbf{0.39} \\
  ResNet-50$^\ast$(95\%) / I1k & Top-1 Acc (\%) & 0.5\% & $\mathbf{\times 41.5}$ & 75.72 & 76.14 & \textbf{0.42}\\
  YOLO$^\ast$(75\%) / COCO & mAP@.5 & 0.3\% & $\mathbf{\times 16.48}$ & 52.6 & 53.5 & \textbf{0.9}\\
  DistilBERT$^\ast$(58\%) / SQuAD & F1 & 0.025\% & $\mathbf{\times 19.4}$ & 84.70 & 84.92 & \textbf{0.22} \\
  \specialrule{.2em}{.1em}{.1em} `
  \end{tabular}
  \normalsize
\end{table*}
\vspace{-\baselineskip}

%% file: sections/06_conclusion.tex
\section{Conclusion}
In this paper, we have investigated a post-training quantization method that strives to minimize the rate of the model's parameters subject to a deviation constraint. A \emph{rotation-invariant quantization} scheme (RIQ) was introduced, which quantizes each layer in proportion to the layer's norm, searching for the optimal solution over the family of spherical distributions. To find the solution efficiently, we derived the scale in which the rate increases with the
deviation and then suggest a searching paradigm that bounds the search space based on our findings. The rate-distortion curve was thoroughly analyzed, showing that the minimizing distribution is a product distribution, constructed as the product of the layer's spherical distribution, which coincides with the RIQ approach.    



%% file: sections/10_appendix.tex
\section{Appendix}
In this section, we provide rigorous proofs for the theorems and the statements herein. Further, we present additional  results for RIQ. 
\ifshort
\subsection{Proof of \texorpdfstring{Lemma~\ref{coro: delta to eps}}{}}\label{sec. proof of coro1}

\input{sections/proofs/coro1}
\subsection{Proof of \texorpdfstring{Corollary~\ref{coro: distortion to parameters distortion}}{}} \label{sec. proof of coro2}
\input{sections/proofs/coro2}

\subsection{Proof of \texorpdfstring{Proposition~\ref{prop: monotonically decreasing}}{}} \label{sec. proof of prop3}
\input{sections/proofs/prop3}

\subsection{Proof of \texorpdfstring{Proposition~\ref{prop. k bounds}}{}} \label{sec. proof of prop2}

\input{sections/proofs/prop2}
\subsection{Proof of \texorpdfstring{Proposition~\ref{prop: rotation invariance of rvwh}}{}} \label{sec. proof of prop1}

\input{sections/proofs/prop1}
\else\fi




\subsection{Relation to Other Error Criteria}\label{sec. relation to other errors}
\begin{remark}
The proof of Lemma~\ref{coro: delta to eps} in \Cref{sec. proof of coro1} may serve as a proxy to other error criteria such as the 
\emph{Signal to Quantization Noise Ratio} (SQNR), \cite{caffarena2010sqnr}. Specifically, similar to the proof of  Lemma~\ref{coro: delta to eps}, the resulting connection between the scale $\Delta_\ell$ and the SQNR $\eps^\prime_\ell$ in each layer $\ell$ is 
\begin{equation*}
  \eps_{\ell}^{\prime} \triangleq \frac{\Vert  \ww_\ell - \hat{ \ww}_\ell  \Vert }{\Vert \ww_\ell \Vert } = \sqrt{\frac{\Delta^2_\ell}{12} \cdot \frac{n_\ell}{\Vert \ww_\ell\Vert^2}}  
\end{equation*}
Or, equivalently, 
\begin{equation*}
  \Delta_\ell = \eps_{\ell}^\prime \Vert \ww_\ell \Vert \sqrt{12/ n_\ell} 
\end{equation*}
\end{remark}

\subsection{Additional Results}\label{sec. additional results}
\subsubsection{Decomposing the Rate-Distortion Curve}
The key steps of lossy compression are quantization and compression. In the quantization phase, the RIQ approach is minimizing the overall model's entropy by allocating a small number of unique symbols for large-norm layers. To achieve (asymptotically) this entropy limit, we utilize the ANS (lossless) entropy encoder. In this section, we evaluate the contribution of each step to the rate-distortion tradeoff. Namely, the average rate per (quantized) symbol before and after ANS. At run-time, when a certain layer is required, it is decoded and represented at a rate according to RIQ. If this rate is below $8$ bits/symbol, it enables significant acceleration by performing $8$ bits integer operations, as discussed in Appendix~\ref{sec. fq}.

\Cref{fig:rate_distortion_resnet} depicts the rate-distortion curve for ResNet-50, decomposed to the quantization step (dashed lines) and the resulting compression step, following the quantization step (solid lines). As a baseline, the uniform scalar quantization (red color) is given for comparison with RIQ (green color). Interestingly, RIQ (dashed green line) outperforms the uniform quantization (dashed red line) by about $\sim 4$ bits/symbol and even its resulting compressed size by about $\sim 1$ bit/symbol. Indeed, the latter indicates that uniform quantization does not minimize the model's entropy. Applying the ANS compression following RIQ reduces additional  $\sim 3$ bits/symbol (solid green line), which according to our analysis is the minimum entropy possible for a given distortion. Moreover, our method achieves a reduction of about $\sim 8$ bits/symbol compared to uniform scalar quantization alone, and an additional $\sim 4$ bits/symbol when ANS is applied to the uniformly quantized weights. For completeness, in \Cref{fig:quant_statistics}, we depict the rate per layer statistics for the ResNet-50 model with a deviation constraint of $0.5\%$, using $\eps_0=0.01$. Noticeably, most of the layers require less than 8 bits. Moreover, the average rate is 6.9 bits per symbol, as larger layers get quantized more aggressively.  Hence, those layers' run time can be accelerated significantly when their input is quantized as well. Yet, if one wishes to enforce a rate that is less than 8 bit to all layers, it should pick an $\eps_0$ according to Remark~\ref{remark:rate enforcement}. 

The rate-distortion curves for various models, in particular, the  VGG (green circles), ResNet-50 (red squares), ViT (blue diamonds), and DistilBERT (orange triangles) are given in \Cref{fig:rate_distorion}. As expected, the curves decrease monotonously, reaching an impressive compression rate of less than $8$ bits/symbol on average even for extremely low cosine distance in all presented models. \Cref{table:more_models} depicts the compression ratio attained for various  models\footnote{\href{https://github.com/onnx/models}{\nolinkurl{github.com/onnx/models}}} and tasks, with  $1\%$ deviation constraint. Interestingly, from this table, we can infer that the MobileNet-v2 and ArcFace models are relatively efficient since their compression ratio is lower than the other (over-parameterized) models.

One of the strengths of RIQ is its efficiency. In fact, for ResNet-50 (whose size is 102 MB), it takes less than 1 minute to find $k$ and compress the model on a CPU. Note, however, that finding the optimal $k$ is typically an offline procedure, and thus it is not time critical.

\begin{figure*}[]
\centering
 \subfigure[]{
 \centering
    \includegraphics[width=0.3\columnwidth]{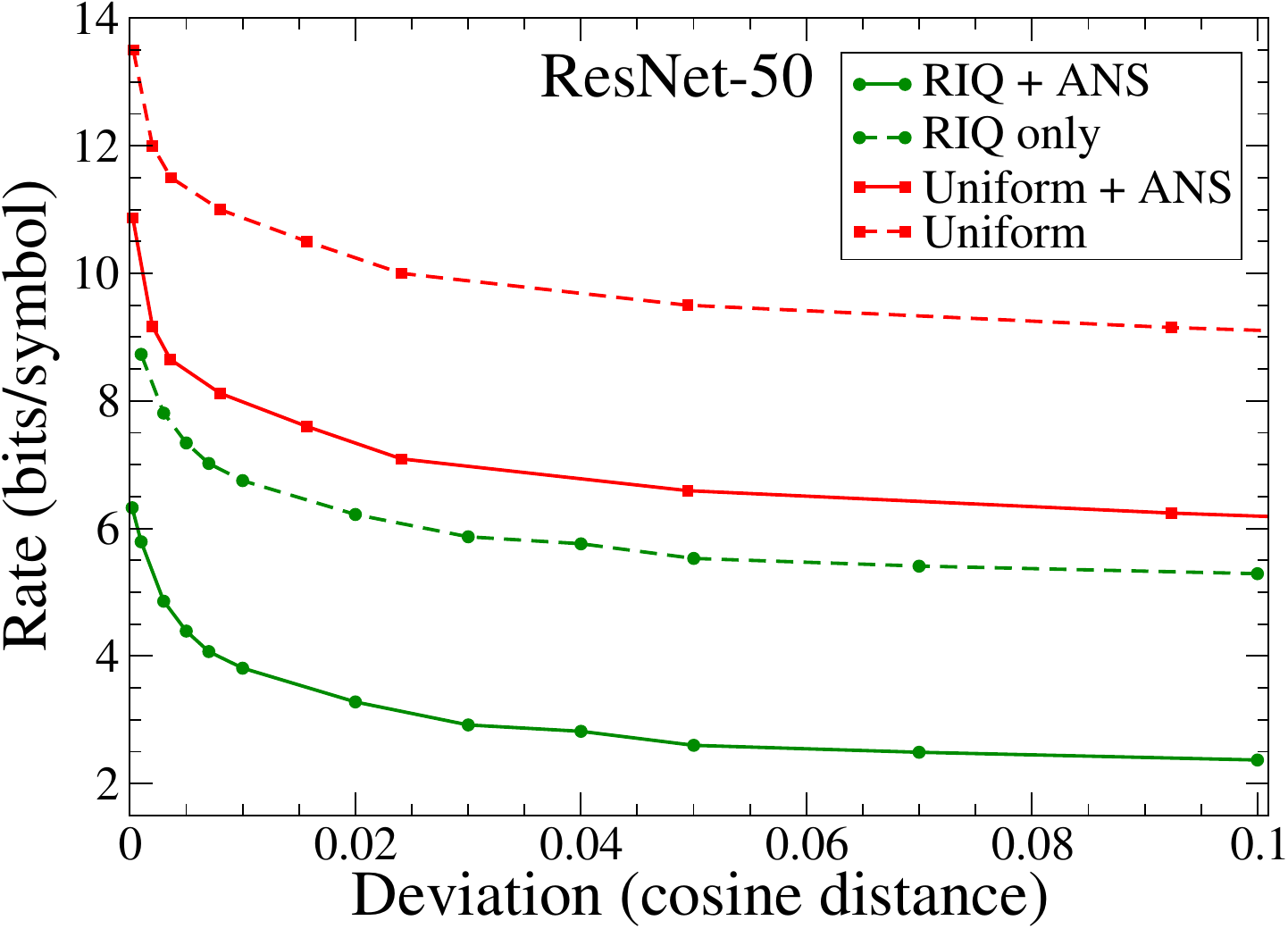} 
    \label{fig:rate_distortion_resnet}
 }
 \hfill
 \subfigure[]{
 \centering
    \includegraphics[width=0.3\columnwidth]{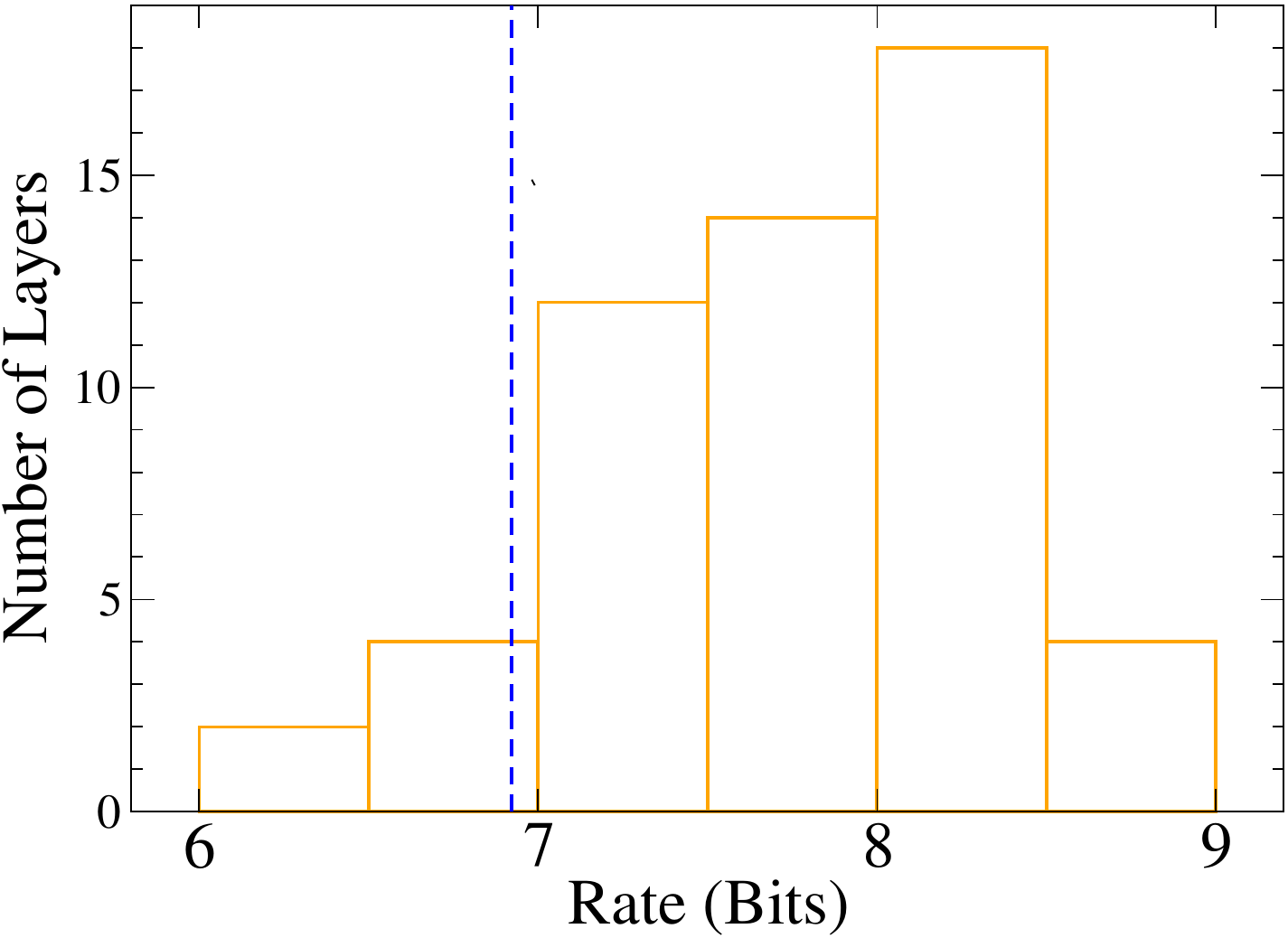} 
    \label{fig:quant_statistics}
 }
 \hfill
 \subfigure[]{
 \centering
    \includegraphics[width=0.3\columnwidth]{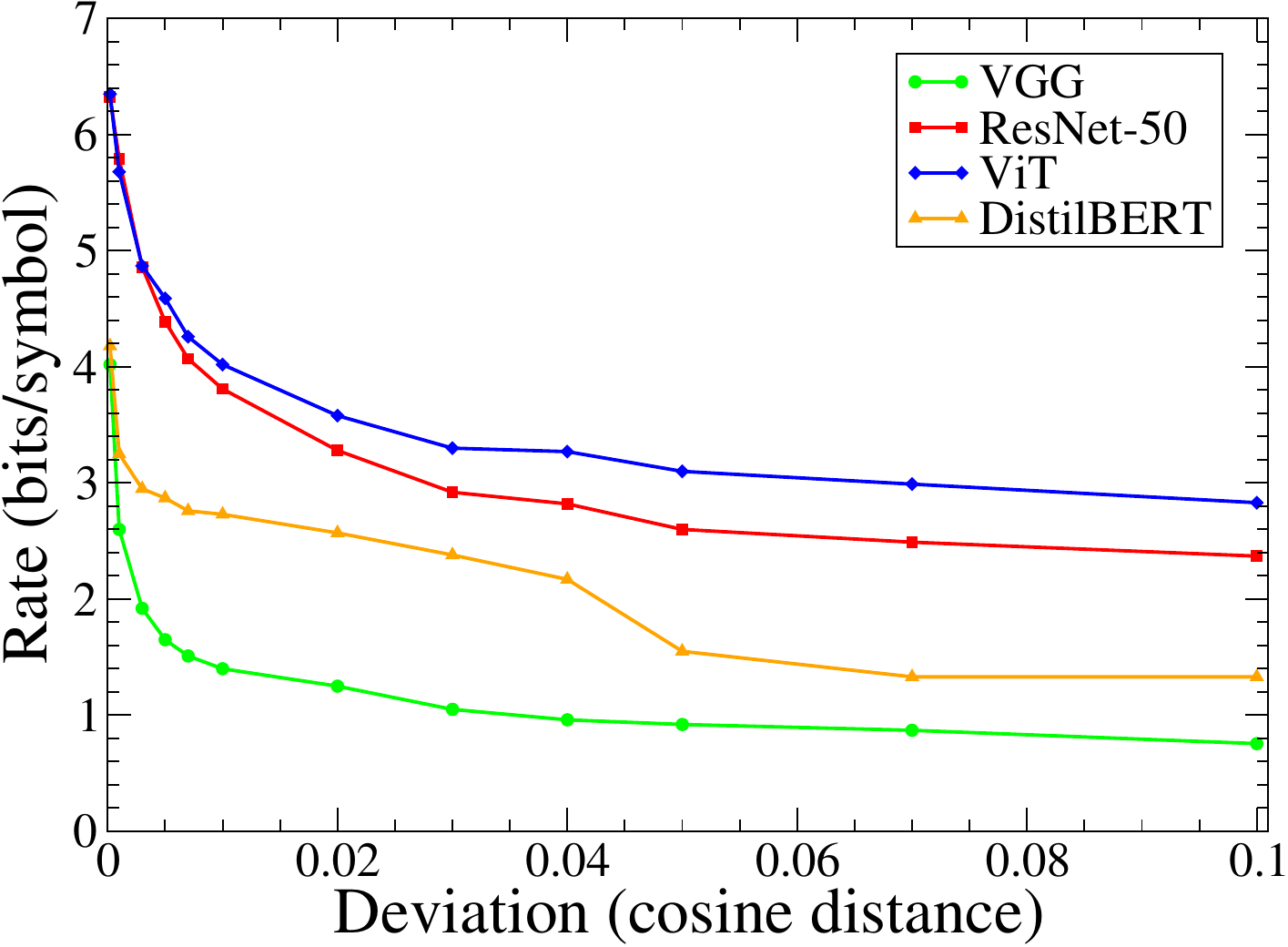}
    \label{fig:rate_distorion}
 }
 \caption{(a) Rate-distortion curve for ResNet-50 model obtained for RIQ (green circles) as well as Uniform linear quantization (red squares). Rates are presented for both the quantized model (dashed) as well as following an ANS compression. (b) ResNet-50 rate per layer statistics with $\eps_0 = 0.01$ in all layers. (c) Rate distortion curves obtained by RIQ + ANS, for a variety of models: VGG (green circles), ResNet-50 (red squares), ViT (blue diamonds), and DistilBERT (orange triangles).}
\end{figure*}

\begin{table*}[ht]
\footnotesize
  \centering 
  \caption{Compression and deviation attained by RIQ for various models and tasks under deviation constraint of 1\%.
  }
  \label{table:more_models}
  \begin{tabular}{ccc}
  \specialrule{.2em}{.1em}{.1em} 
  Model & Deviation & Compression \\
       & constraint & ratio \\
  \specialrule{.2em}{.1em}{.1em} 
  MobileNet-v2 (image classification)  & 1 \% & $\times 5.79$ \\
  ArcFace (face detection)  & 1 \% & $\times7.75$ \\
  AlexNet (image classification) & 1 \% & $\times 19.3$ \\
  Candy (style transfer) & 1 \% & $\times 20.8$ \\
  AgeGoogleNet (gender and age) & 1 \% & $\times 26.95$\\
  \specialrule{.2em}{.1em}{.1em} 
  \end{tabular}
  \normalsize
\end{table*}

\Cref{table:llama-7b} describes the performance of RIQ on large language models. In particular, we used RIQ to compress the llama-7b $\times 4$, and then evaluated the quantized model using \cite{eval-harness}.

\begin{table*}[ht]
\footnotesize
  \centering 
  \caption{Llama-7b performance with $\times 4$ compression. 
  }
  \label{table:llama-7b}
  \begin{tabular}{cccc}
  \specialrule{.2em}{.1em}{.1em} 
  Task & Metric & Baseline & RIQ \\
  \specialrule{.2em}{.1em}{.1em} 
  ARC challenge  & acc (\%) & 41.89 \% & 41.98\% \\
  SWAG  & acc norm  (\%) & 76.60\% & 75.95\% \\
  XWinograd & acc (\%) & 87.87 \% & 87.18\% \\
  piqa & acc (\%) & 78.67\% & 77.97\% \\  
  \specialrule{.2em}{.1em}{.1em}
  \end{tabular}
  \normalsize
\end{table*}

\Cref{table:comparison_various_tasks} compares the accuracy and compression ratio attained by RIQ with various baseline techniques. This tables shows that creating a smaller model by knowledge distillation attains impressive accuracy (sometimes even higher than the baseline), yet, falls short in terms of compression ratio. Other baselines that quantizes both the activations and weights suffer from high accuracy drop.   

\begin{table*}[ht]
  \centering 
  \small
  \begin{threeparttable}
    \caption{Comparison of RIQ with various baseline compression techniques on various tasks.}
  \label{table:comparison_various_tasks}
    \begin{tabular}{c|l|c c c c }
    \specialrule{.1em}{.1em}{.1em} 
    Model & {\small Method} & Comp. & Acc. (\%) & Ref. (\%) & Drop (\%)\\
    \specialrule{.2em}{.1em}{.1em}
    \multirow{2}{*}{ViT I1k} 
    & \small MiniViT (KD) \cite{zhang2022minivit} \kern-1em & {$\times 2$} & \textbf{84.7} & \textbf{77.9} &  \textbf{-6.8}  \\
    & \textbf{RIQ (Ours)} & $\times$ \textbf{ 6.98} & 81.0 & 81.07 & 0.07 \\
    \specialrule{.1em}{.1em}{.1em}
    \multirow{2}{*}{YOLO/COCO} 
     & DRGS  \cite{wu2022drgs} \kern-1em & $\times 8$ & 33.4 & 55.7 & 22.3   \\
     & \textbf{RIQ (Ours)} & $\times$\textbf{8.34} &  \textbf{54.7} & \textbf{55.7} &  \textbf{1.0} \\     
     \specialrule{.1em}{.1em}{.1em}
  \multirow{2}{*}{DistilBERT/SQuAD} 
     & OFA (KD) \cite{zafrir2021prune} \kern-1em & $\times 6.67$ & \textbf{88.82} & \textbf{85.8} & \textbf{-0.02}   \\
     & \textbf{RIQ (Ours)} & $\times$ \textbf{7.96} &  85.0 & 85.08 &  0.08 \\     
    \specialrule{.2em}{.1em}{.1em} 
    \end{tabular}
  \end{threeparttable}
  \normalsize
\end{table*}

\subsubsection{Rotation Invariant Quantization with Quantized Activations}\label{sec. fq}
Quantizing both the NN model's weights and its activations can further accelerate the inference, \cite{wu2020integer, nagel2020up, Krishnamoorthi2018QuantizingDC}. Nevertheless, in this case, the quantization error of both the weights and the activation affects the model's output. In the seminal work of \cite{wu2020integer}, the authors utilized the KL distance for quantizing the activations to minimize the information loss at the output. In this section, we examine the RIQ approach, combining it with activation quantization. 

To demonstrate, we use the \cite{nvidiaquant} quantization library for the ResNet-50 model with a “mini ImageNet” validation set, which comprises one image per class and a total of $1000$ images.  We evaluate this library's performance as a baseline, where the  activations are quantized by the KL-distance criterion, and the weights are quantized to 8-bit linearly. The resulting cosine distance at the output of this baseline is $0.69\%$.  For comparison, this reference value is given as the deviation requirement to RIQ. In particular, to integrate RIQ, the activations are quantized as the baseline, and then, we run RIQ according to Algorithm~\ref{alg: rotation inv}. This way, RIQ is aware of  the activations' quantization error during its search for the single-letter solution. Note that to facilitate the acceleration of int8 operations, RIQ must yield a quantization rate of up to $8$ bits/symbol. In case a certain layer requires a higher rate, we simply perform linear uniform quantization to $8$ bits (without clipping), as the baseline does.  

\begin{figure*}[h]
\centering
 \subfigure[]{
 \centering
    \includegraphics[width=0.42\columnwidth]{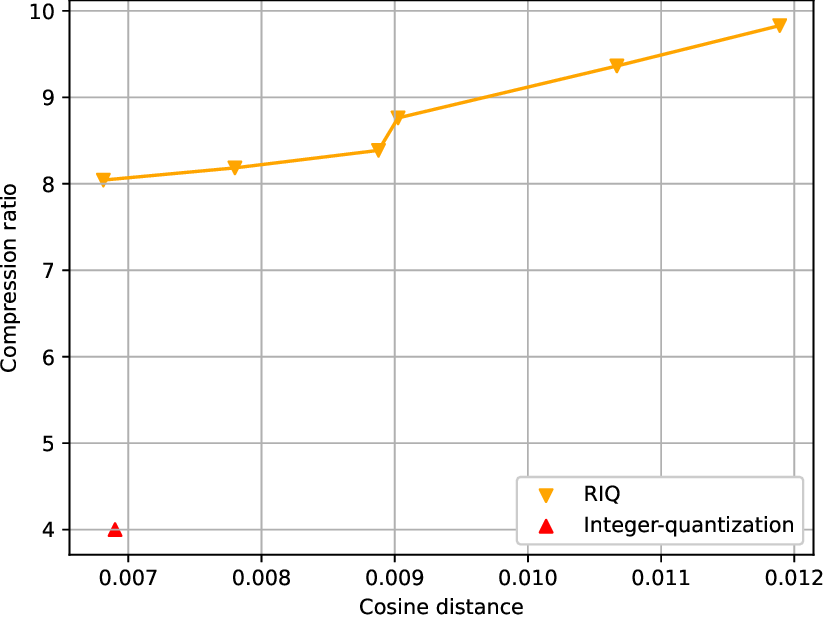}
    \label{fig: RN-50 FQ}
 }
 \hfill
 \subfigure[]{
 \centering
     \includegraphics[width=0.44\columnwidth]{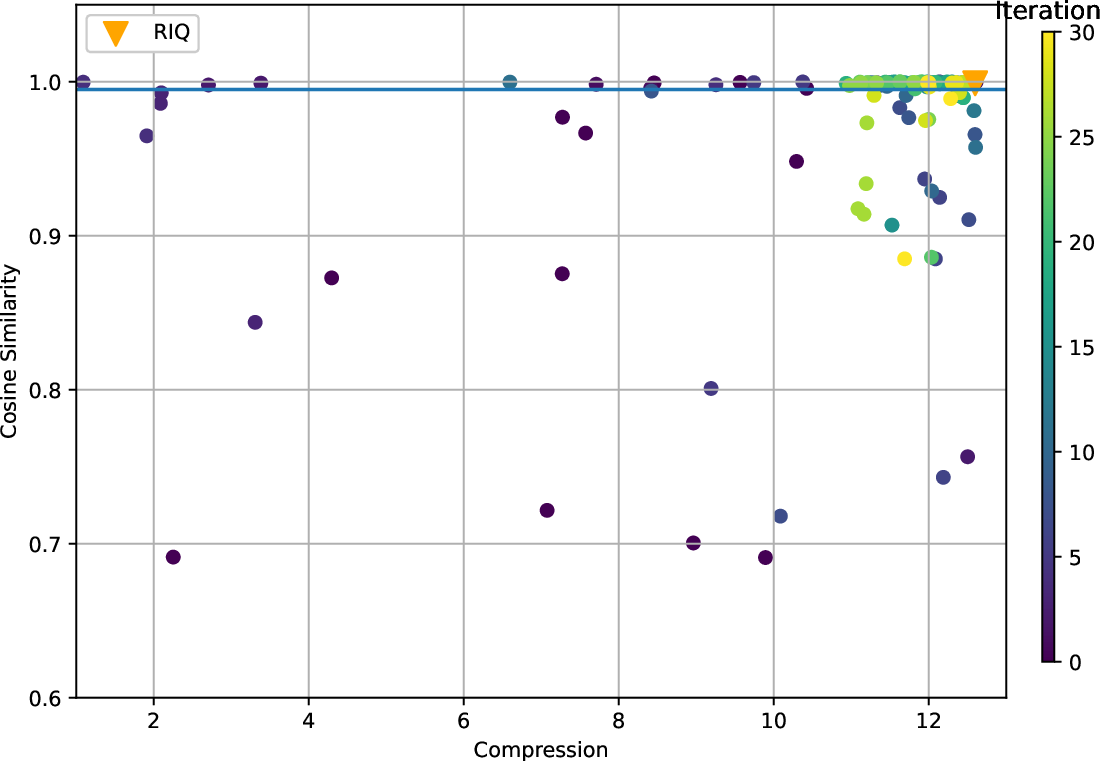}
     \label{fig: mobo_judge}
 }
 \caption{(a) The compression ratio as a function of cosine distance.  The left-bottom red triangle depicts the resulting distance of 0.0069 achieved by the baseline with a compression ratio of $\times 4$. The orange upside-down triangles depict the cosine distance and compression ratio attained by RIQ with the ANS compression. The orange line depicts the trend line. (b) MOBO optimization process. Interestingly, MOBO converges at the few last iterations to $\times 12$ compression, with a highest value of $\times 12.61$. On the other hand, RIQ reaches practically the same compression ratio in a few seconds.}
\end{figure*}

\Cref{fig: RN-50 FQ} characterizes the compression ratio as a function of cosine distance.  The leftmost point reflects a cosine distance of $0.0069$ achieved by the baseline of \cite{nvidiaquant}. Remarkably, the RIQ attains superior compression with relatively low deviation even when the activations are quantized. In run-time, of course, the reconstructed values are represented again by 8-bit value, and hence, the significant acceleration of \cite{wu2020integer} is still valid.

\subsubsection{Comparison with Multi-Objective Bayesian Optimization}
In this section, we utilize the \emph{Multi-Objective Bayesian Optimization} (MOBO) tool, described in \cite{daulton2020differentiable} to compress NN models, and compare results with RIQ. To compress models with MOBO, we set two objective functions for it. The first objective is minimizing the cosine distance in \cref{eq: cosine dist def}. The second objective is maximizing the compression ratio in \cref{eq:cmp_ratio}. Then, we let MOBO optimize the rate-distortion tradeoff (i.e., the Pareto frontier surface).

Nonetheless, MOBO is quite complex and requires strong computing capabilities for exploration and exploitation. Particularly, reaching the optimal solution may take days and even weeks, using multiple GPUs.  Even on small NN models, to address the high-dimensional search spaces, we apply sparse axis-aligned subspace priors for Bayesian optimization (qNEHVI + SAASBO), with the batch Noisy Expected Improvement (qNEI) acquisition function, as suggested by   \cite{Eriksson2021HighDimensionalBO, daulton2021multi, Daulton2021ParallelBO}. Moreover, since the two objectives are not within the same range the cosine similarity objective had to be scaled accordingly to converge to the optimal solution, where a calibration set of 4 images are used during 30 iterations of exploration/exploitation.

Accordingly, we pick a (relatively) small model for comparison (with a size of 112 KB), letting
MOBO to find for each layer its optimal bin width and quantize accordingly. We emphasize that the MOBO solution does not rely on the rotation invariant insights. In \Cref{fig: mobo_judge}, the optimization process of MOBO is presented, where each dot depicts experiment results, and its color indicates the iteration in which this result was attained. The compression results of RIQ are presented for comparison. Remarkably, RIQ and MOBO attained almost identical results of $\times 12.6$ and $\times 12.61$, respectively, with a cosine distance of $0.005$. This indicates that RIQ reaches the optimal solution.


%% file: sections/proofs/coro2.tex
\begin{proof}
Let $\theta_{[1:L]}$ be the rotation angle from $\rvw_{[1:L]}$ to $\rvwh_{[1:L]}$ such that  $\ninp{\rvw_{[1:L]}}{\rvwh_{[1:L]}} \triangleq \cos(\theta_{[1:L]})$. Considering the high-rate regime, where $\Vert \rvwh_\ell \Vert = \Vert \rvw_\ell \Vert + o(\Vert \rvw_\ell \Vert)$, then, 
\begin{align*}
    \cos(\theta_{[1:L]}) & = \ninp{\rvw_{[1:L]}}{\rvwh_{[1:L]}} & \\
    & = \frac{\sum_{\ell=1}^L \inp{\rvw_\ell}{\rvwh_\ell}}{\Vert \rvw_{[1:L]} \Vert \cdot \Vert \rvwh_{[1:L]} \Vert} & \\
    & =  \frac{\sum_{\ell=1}^L\Vert \rvw_{\ell} \Vert \cdot \Vert \rvwh_{\ell} \Vert \cdot \ninp{\rvw_\ell}{\rvwh_\ell}}{\Vert \rvw_{[1:L]} \Vert \cdot \Vert \rvwh_{[1:L]} \Vert} & \\
    & = \sum_{\ell = 1}^L \frac{\Vert \rvw_{\ell} \Vert^2}{\Vert \rvwh_{[1:L]} \Vert^2} \cos(\theta_\ell) + o( \frac{\Vert \rvw_{\ell} \Vert^2}{\Vert \rvwh_{[1:L]} \Vert^2} )
\end{align*}
Since $\sum_{\ell = 1}^L \frac{\Vert \rvw_{\ell} \Vert^2}{\Vert \rvw_{[1:L]} \Vert^2} = 1$, the parameters' distortion is simply a convex combination of the layers' distortion. 
\end{proof}

%% file: sections/proofs/prop3.tex
\begin{proof}
By \cref{eq: lemma 1}, 
\begin{equation*}
      1- \cos(\theta_f)  =  \frac{\Vert \ghM(x) - \gM(x) \Vert^2}{2 \cdot\Vert \gM(x) \Vert^2} + o(1).
\end{equation*}
where $\theta_f$ is the deviation angle obtained by \cref{eq: cosine dist def}.

To make the dependence of $\gM(x)$ on its weights explicit, let us denote $\gM(x) = \gM_x (\rvw_{[1:L]})$ and $\ghM(x) = \gM_x (\rvwh_{[1:L]})$. 
Since the denominator is independent of the quantization, it is sufficient to focus on the enumerator. 
Accordingly, we wish to examine 
\begin{equation*}
    \Vert \gM_x (\rvwh_{[1:L]}) - \gM_x (\rvw_{[1:L]}) \Vert^2  
    =   \Vert \gM_x (\rvw_{[1:L]} + \rveps) - \gM_x (\rvw_{[1:L]}) \Vert^2  
    \triangleq g(\rveps)
\end{equation*}
where $\rveps$ is (random) quantization errors. Our goal is to prove that $\E_\rveps  \brkt{ g(\rveps)}$ is monotonically decreasing in $k$ (i.e., higher rate must reduce the quantized model deviation).  
Using Taylor expansion at $\rveps = \pmb{0}$, we have 
\begin{align*}
     g(\rveps) &=   g(\pmb{0}) + \nabla g(\pmb{0}) \cdot \rveps + \frac{1}{2} \rveps^T \gH(g(\pmb{0})) \rveps  + o(\rveps^3) &\\
    & =  \frac{1}{2} \rveps^T \gH(g(\pmb{0})) \rveps   &\\
\end{align*}
where the last step follows since $g(\pmb{0}) = \pmb{0}$, and noting that $\nabla g (\pmb{0}) = 0$. Finally, omitting the little order $o(\rveps^3)$, which is negligible in the high rate regime.

Recall that in the high rate regime the error in entry $i$, $\rveps_i \sim U[-\Delta/2,\Delta/2]$ is i.i.d.\ uniformly distributed.  Accordingly, in each layer $\ell$ the corresponding sub-vector $\rveps_\ell$ satisfies $\E\brkt{\rveps_\ell \rveps_\ell^T} = \Delta_\ell^2 / 12 \cdot \rmI_{n_\ell}$, where $\rmI_{n_\ell}$ is the $n_\ell\times n_\ell$ identity matrix. Since the errors are independent with zero mean and variance $\Delta_\ell^2 / 12$, we can utilize the \citet{Hutchinson1989ASE} trick. Accordingly, let $N = \sum_{\ell=1}^L n_\ell$, then
\begin{align*}
    \frac{1}{2} \E_\rveps \brkt{  \rveps^T \gH(g(\pmb{0})) \rveps } & = \frac{1}{2} \E_\rveps \brkt{ \sum_{i=1,j=1}^{N}  \eps_i  \gH(g(\pmb{0}))_{ij} \eps_j } &\\
    & =  \sum_{i=1,j=1}^{N}  \gH(g(\pmb{0}))_{ij} \E_\rveps \brkt{ \eps_i \eps_j } & \\
    & = \sum_{i=1}^{N}  \gH(g(\pmb{0}))_{ii} \E_\rveps \brkt{ \eps_i^2 } & \\
\end{align*}
where the last step follows since $\E_\rveps[\eps_i \cdot \eps_j] = 0 \forall i\neq j$. Finally, letting 
$$\vec{\Delta} \triangleq \prnt{\frac{\Delta_1^2}{12}, \dots, \frac{\Delta_L^2}{12} }^T  =  \frac{1}{k^2}\prnt{\frac{\Vert \rvw_1 \Vert^2}{12}, \dots, \frac{\Vert \rvw_L \Vert^2}{12}}^T, $$
we can present the last step vectorially as 
$$\vec{\Delta}^T \mathrm{diag}\prnt{ \gH(g(\pmb{0}))} = \frac{1}{12 \cdot k^2} \cdot \prnt{\Vert \rvw_1 \Vert^2, \dots, \Vert \rvw_L \Vert^2} \cdot  \mathrm{diag}\prnt{ \gH(g(\pmb{0}))} .$$
Since $g(\pmb{0})$ is a quadratic function with minimum at $\rveps = \pmb(0)$, thus $ \gH(g(\pmb{0}))$ is positive definite, which means that its diagonal entries are real and non-negative. Further, the entries of $\vec{\Delta}$ are positive, hence, this dot product is monotonically decreasing in $k$, as $O(1/k^2)$, which completes the proof.  
Note that this proposition holds for any $\rvx$.
\end{proof}

\begin{figure*}[h]
\centering
 \subfigure[]{
 \centering
   \includegraphics[width=0.435\textwidth]{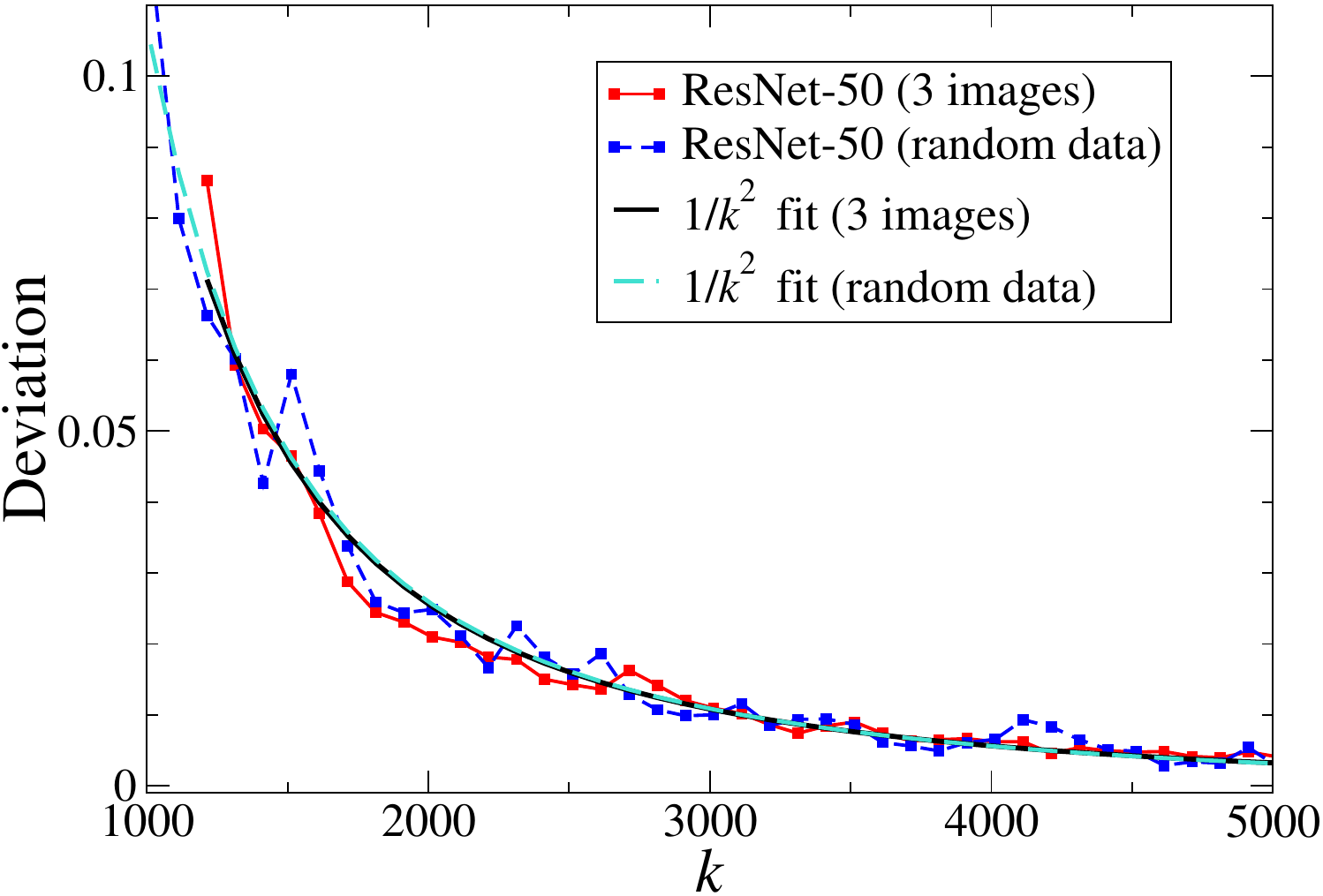}
     \label{fig: validation prop}
 }
 \hfill
 \subfigure[]{
 \centering
     \includegraphics[width=0.505\columnwidth]{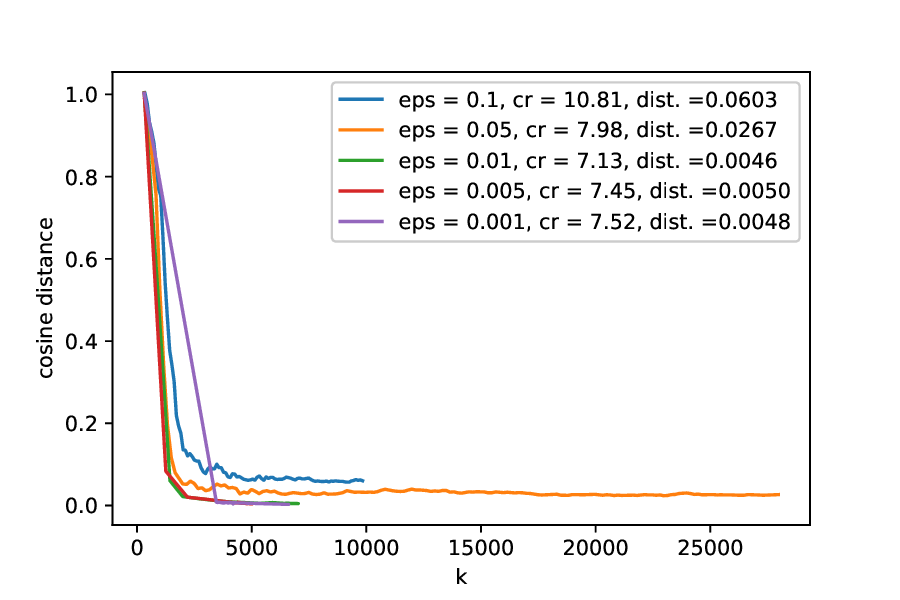}
     \label{fig: eps_0 influence}
 }
 \caption{(a) Validation of Proposition~\ref{prop: monotonically decreasing} on ResNet-50. (b) The impact of $\eps_0$ on the performance of RIQ on ResNet50. Higher values of $\eps_0$ attains higher compression, yet, higher deviation.}
\end{figure*}

 \Cref{fig: validation prop} validates the analysis of Proposition~\ref{prop: monotonically decreasing}. In particular,  we measure the deviation for ResNet-50 as a function of $k$ using both random data and a calibration set. Then, we compare the resulting deviations with a fitted curve of the form $y=a/k^2$. Clearly, the monotonicity allows to search efficiently the optimal solution.

